%% file: root.tex
\documentclass[letterpaper, 10 pt, conference]{ieeeconf}  

\IEEEoverridecommandlockouts                              

\title{\LARGE \bf
Rectify, Don’t Regret: On-Policy Closed-Loop Training for Multimodal Trajectory Prediction
}

\author{Harsh Yadav, Christian Bohn, and Tobias Meisen \\
\textit{University of Wuppertal,} harsh.yadav@uni-wuppertal.de
\vspace{-3mm}
\thanks{This work has been submitted to the IEEE for possible publication. Copyright may be transferred without notice.}
}

\usepackage{threeparttable}
\usepackage[dvipsnames]{xcolor}
\definecolor{cvprblue}{rgb}{0.21,0.49,0.74}
\usepackage[pagebackref=false,breaklinks,colorlinks,allcolors=cvprblue]{hyperref}
\usepackage{amsmath}
\usepackage{subcaption}
\usepackage{multirow}
\usepackage{hhline}
\usepackage{tikz}
\usepackage{amssymb}
\usepackage{float}
\usepackage[ruled,vlined]{algorithm2e}

\SetCommentSty{mycommentfont}
\DeclareMathOperator*{\argmin}{argmin}
\usepackage{censor}
\usepackage{acro}
\usepackage{siunitx}
\usepackage{acro}
\usepackage[noadjust]{cite}
\DeclareAcronym{wta}{
	short=WTA,
	long=Winner-Takes-All,
}
\DeclareAcronym{nll}{
	short=NLL,
	long=Negative Log Likelihood,
}
\DeclareAcronym{obb}{
	short=OBB,
	long=Oriented Bounding Box,
}
\DeclareAcronym{ood}{
	short=OOD,
	long=out-of-distribution,
}
\DeclareAcronym{mpc}{
	short=MPC,
	long=Model-Predictive-Control,
}
\DeclareAcronym{ade}{
	short=ADE,
	long=Average-Displacement-Error,
}
\DeclareAcronym{gnn}{
	short=GNN,
	long=Graph Neural Network,
}
\DeclareAcronym{cnn}{
	short=CNN,
	long=Convolution Neural Network,
}
\DeclareAcronym{sota}{
	short=SOTA,
	long=state-of-the-art,
}
\DeclareAcronym{bev}{
	short=BEV,
	long=Bird-Eye-View,
}
\DeclareAcronym{ad}{
	short=AD,
	long=Autonomous Driving,
}
\DeclareAcronym{rl}{
	short=RL,
	long=Reinforcement Learning,
}
\DeclareAcronym{ol}{
	short=OL,
	long=open-loop-trained,
}
\DeclareAcronym{cl}{
	short=CL,
	long=closed-loop-trained,
}
\DeclareAcronym{bptt}{
	short=BPTT,
	long=Backpropagation Through Time,
}
\DeclareAcronym{av2}{
	short=AV2,
	long=Argoverse-2,
}

\begin{document}
\bstctlcite{IEEEexample:BSTcontrol}

\maketitle
\thispagestyle{empty}
\pagestyle{empty}

\input{sec/0_abstract}
\input{sec/1_intro}
\input{sec/2_related_work}
\input{sec/3_methods}
\input{sec/4_experiments}
\input{sec/5_results}

\input{sec/6_conclusion}

\bibliographystyle{IEEEtran}
\bibliography{IEEEabrv,mybibfile}

\end{document}

%% file: sec/0_abstract.tex
\begin{abstract}
Current trajectory prediction models are primarily trained in an open-loop manner, which often leads to covariate shift and compounding errors when deployed in real-world, closed-loop settings. In this work, we propose a closed-loop training paradigm for receding-horizon ego prediction and identify two design choices that are critical for it to succeed. First, closed-loop supervision must be on-policy: each predicted mode is rolled out and supervised only from the states its own execution produced, preserving the prediction accuracy that cross-mode supervision would otherwise degrade. Second, gradients must not flow across the executed state transition, as backpropagating through these transitions allows the model to retroactively adjust an earlier prediction so that the resulting state is easier to predict from, mathematically \textit{regretting} a mistake rather than learning to recover from it. We show empirically that this shortcut couples the learned policy to the specific replanning rate used during training and degrades at other rates, whereas detaching the gradient forces the model to \textit{rectify} drift instead, allowing it to generalize across replanning rates unseen during training. Evaluated across three benchmarks (nuScenes, DeepScenario, and Argoverse 2) and two architectures, a decoder-only adaptation of LMFormer and an unmodified implementation of QCNet, our approach matches open-loop baselines under single-shot evaluation and improves with higher replanning rates, reducing collisions by up to 35\% and off-road rates by up to 67\%. Our code is available at \href{https://github.com/harshy105/CL-Traj/}{github.com/harshy105/CL-Traj/}.
\end{abstract}

%% file: sec/1_intro.tex
\section{Introduction}
\label{cl_paper_sec:introduction}

Current trajectory prediction models \cite{nayakanti2022wayformer, zhou2023query, liu2024laformer, yadav2025lmformer, knoche2025donut} are commonly trained and benchmarked in an open-loop setting \cite{caesar2020nuscenes, wilson2023argoverse, ettinger2021large}. In deployment, however, a predictor is queried at every control cycle, and a downstream planner or controller executes only the initial segment of its forecast before the scene is re-observed and replanned \cite{kouvaritakis2016model}. This exposes the model to a covariate shift, as it must continuously predict from states likely corrupted by its own prior errors \cite{dauner2023parting, karkus2025beyond}. While downstream planners can absorb these compounded errors through conservative safety buffers \cite{zheng2026contingency}, this comes at the expense of operational efficiency. Rather than relying on such mitigation, this work focuses on making the prediction model itself more robust under frequent replanning through a closed-loop training framework. In addition to standard open-loop training, we identify two key elements essential to maintain prediction stability during frequent replanning at deployment.

The first element addresses on-policy mode supervision. A closed-loop rollout is inherently unimodal: at any timestep, only one predicted mode can be executed to generate the subsequent state. Therefore, supervising a mode using the training signal from a state generated by a different mode introduces off-policy bias, which unfairly penalizes that mode for errors that it did not cause. To prevent this, we propose a structurally isolated, mode-consistent on-policy setup where closed-loop feedback supervises only the mode that generated the state. As shown in Section \ref{cl_paper_subsec:ablation_on_off_policy}, our on-policy setup helps the model retain prediction accuracy under high replanning frequencies, unlike the off-policy variant.

\begin{figure}
    \centering
    \includegraphics[width=\linewidth]{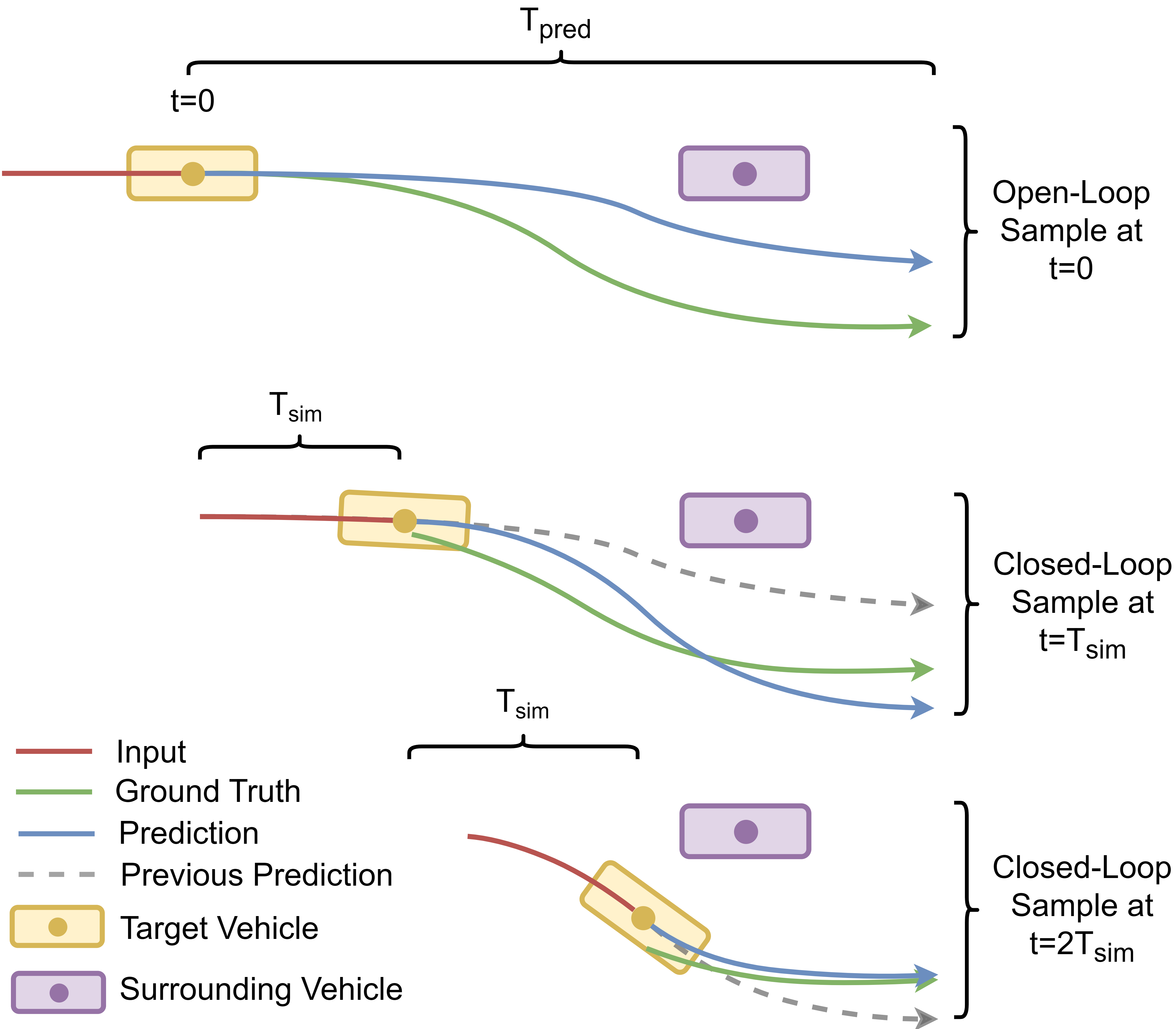}
    \caption{On-policy closed-loop rollout for a single predicted mode. At $t{=}0$ the model produces a full $T_{\text{pred}}$-horizon prediction from the logged state (open-loop sample). The first $T_{\text{sim}}$ seconds are executed, advancing the ego to a drifted state; the model then re-predicts from this self-induced state (closed-loop sample at $t{=}T_{\text{sim}}$). The process repeats at $t{=}2T_{\text{sim}}$. Surrounding agents advance via log replay. Gradients are detached at each state transition (Sec.~\ref{cl_paper_sec:method_grad_detach}), so each planning cycle is supervised independently. Every resulting prediction is supervised against the static ground truth over its remaining horizon (Green).}
    \label{cl_paper_fig:method_cl_simulator}
    \vspace{-5mm}
\end{figure}

The second element concerns whether gradients should flow backward through the state transition, a core mechanism used in \ac{bptt} approaches \cite{werbos1990backpropagation, jaeger2002tutorial}. Because \ac{bptt} is foundational for learning over sequential data \cite{jaeger2002tutorial}, several recent works in autonomous driving have adopted it by making the closed-loop rollout fully differentiable across timesteps \cite{suo2021trafficsim,scheel2022urban,gulino2023waymax,lin2025revisit}. The implicit assumption is that gradient flow across time would yield better policies, mirroring classical sequence modeling \cite{sutskever2014sequence,tallec2017unbiasing}. However, we identify that, in the case of trajectory prediction, \ac{bptt} can lead to the issue of frequency coupling, as shown in Section \ref{cl_paper_subsec:ablation_diff}. During evaluation, a model trained with \ac{bptt} attains its lowest collision rate precisely at the replanning period used during training, but degrades at other frequencies. Conversely, we find that stopping gradients across state transitions severs this coupling, enabling the model to generalize better across higher evaluation replanning rates unseen during training. We hypothesize that allowing gradients to flow through executed transitions enables the model to retroactively alter past predictions to land in easier-to-predict states, effectively \textit{regretting} past errors rather than learning to recover from them. Detaching these gradients forces the model to actively \textit{rectify} drift from its current state instead. Consequently, this prevents the policy from coupling to the specific replanning rate used during training. As a practical consequence, this allows a model trained at a low, computationally cheap replanning rate to generalize better to higher replanning frequencies at deployment.

We validate our closed-loop training setup across three benchmarks
\cite{caesar2020nuscenes, lu2023deepscenario, wilson2023argoverse} and two architectures: a decoder-only variant of LMFormer \cite{yadav2025lmformer} and an unmodified QCNet \cite{zhou2023query}, confirming that the effect belongs to the training paradigm rather than a single model. Our contributions are:

\begin{itemize}
    \item An on-policy closed-loop training paradigm for multi-modal trajectory prediction. The resulting model matches open-loop training under single-shot open-loop evaluation and improves with the replanning rate, reducing collisions by up to $35\%$ and off-road rate by up to $67\%$. 

    \item We isolate key factors governing performance, showing that (i) on-policy rollouts preserve prediction accuracy at high replanning rates, (ii) gradient detachment reduces frequency coupling and improves generalization across higher replanning rates, and (iii) the ego-agent achieves similar performance using log-replay for surrounding agents compared to a goal-conditioned scene prediction model, but at lower computational costs.

    \item Following the decoder-only trend in recent prediction architectures \cite{knoche2025donut}, a decoder-only backbone, LMFormer-D, with $45\%$ fewer parameters and $62\%$ fewer floating point operations at matched prediction quality to LMFormer \cite{yadav2025lmformer}, is presented as the design choice that makes closed-loop rollout under frequent replanning affordable to train.
\end{itemize}

%% file: sec/2_related_work.tex
\section{Related Work}
\label{sec:related_work_cl_paper}

\subsection{Open-Loop Trajectory Prediction}
To address the inherent ambiguity of human behavior, open-loop trajectory prediction has largely progressed toward probabilistic, multimodal modeling. Early approaches \cite{cui2019multimodal, chai2019multipath, gilles2021home, yadav2024caspformer} relied on rasterized Bird-Eye-View inputs processed by convolutional neural networks. To capture multimodality, these models typically adopted a Winner-Takes-All strategy, which splits the objective into a classification term over mode probabilities and a regression term applied only to the mode closest to the ground truth. The rise of transformers then shifted the field toward efficient vectorized input representations \cite{liang2020learning, gao2020vectornet, varadarajan2022multipath++, nayakanti2022wayformer, lan2023sept, liu2024laformer}, enabling direct encoding of map polylines and agent states without lossy rendering. A further paradigm shift within vectorized approaches was introduced by GoRela \cite{cui2022gorela} and QCNet \cite{zhou2023query}, which utilized relative position embeddings to achieve equivariance and eliminate recomputation overhead during autoregressive updates. While subsequent works \cite{knoche2025donut, yadav2025lmformer} achieved \ac{sota} performance using this architecture, recent studies \cite{dauner2023parting, karkus2025beyond} reveal that open-loop trained models fail to translate into closed-loop safety due to covariate shift, where compounding prediction errors cause the ego agent to drift into unfamiliar out-of-distribution states.

\subsection{Closed-loop training and covariate shift}
DAgger~\cite{ross2011related} first addressed the issue of covariate shift by relying on an expert to interactively label actions for unfamiliar states encountered during deployment. However, the need for manual labeling limits the scalability of this approach~\cite{karkus2025beyond}. In contrast, \ac{rl}-based methods such as CaRL~\cite{jaeger2025carl} and GIGAFlow~\cite{cusumano2025robust} completely remove the dependency on expert supervision, instead relying on handcrafted rewards and progression metrics. However, these methods are unstable to tune and often suffer from sim-to-real gaps \cite{li2024platform, karkus2025beyond}. To avoid these pitfalls, some works have pursued closed-loop imitation learning without unstable rewards or manual labeling. DART~\cite{laskey2017dart} and ChauffeurNet~\cite{bansal2018chauffeurnet} perturb the ego pose to generate deviations from the expert trajectory. However, this can lead to states that may not be kinematically feasible. CAT-K \cite{zhang2025closed} improves robustness by fine-tuning the model on its own closed-loop rollouts, but restricts itself to tokenized action spaces. Concurrent to our work, RoaD \cite{garcia2026road} extends CAT-K to continuous and stochastic policies by using expert-guided rollouts as demonstrations. However, both Cat-K and RoaD assume that fine-tuning requires the rollout to stay close to the expert trajectory. In contrast, our framework uses unguided rollouts and supervises drifted predictions directly against the original logged future, within a joint training objective with open- and closed-loop samples rather than a fine-tuning stage.

Finally, inspired by \ac{bptt}, continuous action-space methods like Urban Driver~\cite{scheel2022urban}, TrafficSim~\cite{suo2021trafficsim}, and UniMM~\cite{lin2025revisit} use differentiable simulation to generate closed-loop samples directly within the optimization graph. In modern high-fidelity end-to-end stacks, this is often not realizable. The simulator itself, which may be a renderer, a physics engine, or a high-capacity learned world model, sits between consecutive forward passes. Keeping it in the optimization graph over many steps becomes either non-differentiable or computationally infeasible. The standard remedy is truncated backpropagation through time (TBPTT)~\cite{tallec2017unbiasing}, which confines the backward pass to a limited window, treating the full gradient as the theoretical ideal. However, applying this to closed-loop trajectory prediction risks a distinct failure mode: backpropagating through the executed transition may allow the model to retroactively adjust an earlier prediction so that the resulting state is easier to predict from, effectively regretting a past mistake rather than learning to recover from it. We investigate this hypothesis empirically in Sec.~\ref{cl_paper_subsec:ablation_diff}, and show that the non-differentiable setup generalizes better than the differentiable one across higher replanning rates unseen during training.


\subsection{Closed-loop evaluation}
Standard open-loop benchmarks~\cite{caesar2020nuscenes, wilson2023argoverse, ettinger2021large} score a single prediction against a logged future and cannot observe compounding error. Closed-loop alternatives exist but differ in scope: nuPlan~\cite{nuplan2021Caesar} and CARLA~\cite{dosovitskiy2017carla} evaluate route-conditioned single trajectories, while NAVSIM~\cite{dauner2024navsim} scores pre-computed drift counterfactuals rather than executing a live replanning loop. To our knowledge, no existing benchmark evaluates unconditional multimodal prediction under iterative replanning. We address this gap by building a closed-loop rollout setup that executes ego multimodal predictions alongside recorded logs of surrounding agents, across a sweep of replanning periods.

%% file: sec/3_methods.tex
\section{Method}
\label{cl_paper_sec:method}

Our primary objective is to learn robust ego prediction by leveraging both open-loop data and generated closed-loop rollouts, while strictly maintaining the ego's multimodal behavior. We identify two complementary design choices for robust closed-loop training, as already motivated in Sec.~\ref{cl_paper_sec:introduction}: (i)~a mode-consistent on-policy rollout that preserves prediction accuracy at high replanning rate, and (ii)~gradient detachment across the executed state transition, which prevents the learned policy from coupling to the specific replanning rate used during training. We find each to be independently beneficial in Sections \ref{cl_paper_subsec:ablation_on_off_policy} and \ref{cl_paper_subsec:ablation_diff}, respectively.

\subsection{Mode-Consistent On-Policy Closed-Loop Rollout}
\label{cl_paper_sec:method_on_policy}
A multimodal predictor outputs $M$ candidate trajectories alongside their individual probabilities. The regression head produces the trajectories themselves, while the classification head scores how likely each one is \cite{zhou2022hivt, zhou2023query, knoche2025donut}. In a closed-loop rollout, only one of these candidates can be executed to advance the ego, so the state the model observes next is the product of that single mode. We therefore optimize each predicted mode solely from the states generated by its own execution, avoiding the off-policy bias of supervising a mode from a drifted state it did not produce.

Training thus draws two sample types. An open-loop sample is a prediction made from a logged state, following standard training \cite{zhou2022hivt, zhou2023query}. A closed-loop sample, by contrast, is a prediction made from a state the model itself produced by executing a prior prediction. Consequently, open-loop samples supervise the full multimodal distribution over both classification and regression. Closed-loop samples, being unimodal realizations, supervise only the regression loss of the executed mode, which has no remaining probability left to predict. A computational consequence is that training and evaluation differ in the number of modes rolled out. During training under the winner-takes-all strategy, only the best mode $m^*$ (identified from the initial open-loop sample) is unrolled. Thus, closed-loop overhead scales with $N_{\text{replan}}$ rather than $M \times N_{\text{replan}}$. During evaluation, all $M$ modes are rolled out sequentially, each from its own chain of drifted states, to preserve the on-policy setup at test time. An illustration of the rollout for a single mode is shown in Figure~\ref{cl_paper_fig:method_cl_simulator}, and the complete training procedure is given in Algorithm~\ref{cl_paper_alg:method_cl_training}.

\begin{algorithm}
\caption{On-Policy Closed-Loop Training}
\label{cl_paper_alg:method_cl_training}
\DontPrintSemicolon
\KwIn{Prediction Horizon $T_{pred}$}
\hspace{10mm} Replanning Period $T_{sim}$ \\
\hspace{10mm} Regression ($\theta$) and classification ($\phi$) params

\BlankLine
$N \gets \lfloor (T_{pred}-1)/ T_{sim} \rfloor$  \tcp*[r]{no. of closed-loop samples}

\BlankLine
\Repeat{Convergence}{
    \tcp{Get open-loop sample (upto $t=0$) and ground truths}
    Get static context input, $\mathcal{M}$ 
    \vspace{0.5mm}
    
    Get dynamic context input, $X_0 \in \mathbb{R}^{A \times T_{in} \times 2}$
    \vspace{0.5mm}
    
    Get ego ground truth, $Y_{ego} \in \mathbb{R}^{T_{pred} \times 2}$
    \vspace{0.5mm}
    
    Get surr. agents logs, $Y_{surr} \in \mathbb{R}^{A-1 \times T_{pred} \times 2}$
        
    \BlankLine
    \tcp{simulation rollout to generate $N$ closed-loop samples}
    \For{$n \gets 0$ \KwTo $N$}{
        \tcp{multimodal ego predictions with $m$ modes}
        \vspace{1mm}
         $[(\mu, \Sigma)_{(n,m)}]_{n T_{sim} + 1}^{T_{pred}} \gets \mathcal{F}_{\theta}(\mathcal{M}, X_n)$  
    
        \BlankLine
        \tcp{ego's mode probabilities and the mode to unroll} 
        \If{n==0}{
            $\pi_m \gets \mathcal{F}_{\phi}(\mathcal{M}, X_0)$ 
            
            $m^* \gets \argmin\limits_{m} \sum\limits_{t=1}^{T_{pred}}  || Y_{ego}^t - \mu_{(0,m)}^t ||_{\raisebox{-2pt}{\scriptsize 2}}$ 
        }
        \tcp{update dynamic input with ego pred. \& surr. logs}
        \vspace{1.5mm}
        
        $pos_{ego} = \mathrm{stop\_gradient} \left([\mu_{(n,m^*)}]_{n T_{sim}+1}^{(n+1)T_{sim}} \right)$
        
        $pos_{surr} = [Y_{surr}]_{n T_{sim}+1}^{(n+1)T_{sim}}$
        \vspace{1.5mm}
        
        $X_{n+1} = \left(pos_{ego} || pos_{surr}\right) \in \mathbb{R}^{A \times T_{sim} \times 2}$
        \vspace{1.5mm}

        \tcp{only keep $T_{in}$ number of input frames}
        $X_{n+1} = \left[X_{n+1} \right]_{(n+1)T_{sim}-T_{in}}^{(n+1)T_{sim}} \in \mathbb{R}^{A \times T_{in} \times 2}$             
    }
    \BlankLine
    \tcp{update cls. params. with open-loop sample}
    $\nabla_{\phi} \! \left( \sum\limits_{m=1}^M \pi_m(\phi)  \sum\limits_{t=1}^{T_{pred}}  -\mathrm{log}\left( \mathcal{L}(Y_{ego}^t | (\mu, \Sigma)_{(0,m)}^t(\theta) \right) \! \right)$ \\
    
    \vspace{1mm}
    \tcp{update reg. params. with open- \& closed-loop sample}
    $ \nabla_{\theta} \! \left( \sum\limits_{n=0}^{N} \lambda_n \!\! \sum\limits_{t=n T_{sim}+1}^{T_{pred}} \!\!\!\! -\log\left( \mathcal{L}(Y_{ego}^t | (\mu, \Sigma)_{(n,m^*)}^t(\theta) \right) \! \right)$ \\
    
}
\BlankLine
\end{algorithm}

Our simulation setup only updates the dynamic context, treating map elements as static---a valid approximation over our short prediction horizon ($T_{\text{pred}} = 6\,\text{s}$). At each replanning step, the ego's predicted trajectory is executed for $T_{\text{sim}}$ seconds. This receding-horizon protocol mirrors model predictive control~\cite{kouvaritakis2016model}, where only the initial segment of a planned trajectory is executed before the controller re-observes and replans. The more consequential approximation is that of log-replay: once the ego deviates, surrounding agents continue along their logged paths rather than reacting to it. We attempt to relax this in Sec.~\ref{cl_paper_subsec:ablation_log_vs_scene} by simulating surrounding agents with the same prediction model, but this largely reproduced log-replay trajectories rather than rendering the environment reactive, so we retain the log-replay setup.

\subsection{Gradient Detachment Across State Transitions}
\label{cl_paper_sec:method_grad_detach}

As argued in Sec.~\ref{cl_paper_sec:introduction}, allowing gradients to flow backward through the state transition lets the model regret its past mistakes rather than learning to rectify and recover from them. We therefore detach gradients across state transitions, optimizing each open- and closed-loop sample independently. In Algorithm~\ref{cl_paper_alg:method_cl_training}, the \texttt{stop\_gradient} operator is applied to the ego's predictions before they enter the next cycle's input context. We confirm in Sec.~\ref{cl_paper_subsec:ablation_diff} that a differentiably trained model attains its lowest collision rate precisely at its training replanning period, while the non-differentiable variant generalizes better to smaller replanning periods without retraining.

\subsection{Loss Formulation}
\label{cl_paper_sec:loss}

Following standard practice~\cite{zhou2022hivt, zhou2023query, yadav2025lmformer, knoche2025donut}, we decouple mixture model optimization into classification and regression, but introduce a key structural distinction: closed-loop samples are unimodal realizations and exclusively supervise regression, while only the initial open-loop sample supervises both. The resulting classification and regression losses are defined as follows:
$$
    L_{cls}(\phi) = \sum\limits_{m=1}^M \pi_m(\phi)  \sum\limits_{t=1}^{T_{pred}}  -\mathrm{log}\left( \mathcal{L}(Y_{ego}^t | (\mu, \Sigma)_{(0,m)}^t(\theta) \right)
$$
$$
    L_{reg}(\theta) = \sum\limits_{n=0}^{N} \lambda_n \!\! \sum\limits_{t=n T_{sim}+1}^{T_{pred}} \!\!\!\! -\log\left( \mathcal{L}(Y_{ego}^t | (\mu, \Sigma)_{(n,m^*)}^t(\theta) \right) 
$$
$$
    m^* \gets \argmin\limits_{m} \sum\limits_{t=1}^{T_{pred}}  || Y_{ego}^t - \mu_{(0,m)}^t ||_{\raisebox{-2pt}{\scriptsize 2}}
$$
\noindent where likelihood function $\mathcal{L}$ is adopted from the original open-loop formulations of LMFormer~\cite{yadav2025lmformer} and QCNet~\cite{zhou2023query}, respectively. $(\mu, \Sigma)_{(n,m)}$ represents the mean and variance of the predicted trajectory for $n_{th}$ sample and $m_{th}$ mode. The best mode ($m^*$) is identified via the initial open-loop sample and remains fixed throughout the rollout, preserving the on-policy setup. The weighting factor ($\lambda_n$) balances each sample's contribution within the regression objective. Conceptually, the learning objective shifts across $N$ samples: at $n=0$ (open-loop), the model predicts a full ego trajectory from static samples. For $n \ge 1$ (closed-loop), the ego predictions are generated from drifted states caused by its own prior actions. Optimizing the predictions from these drifted states against the original ground-truth trajectory ($Y_{ego}$) explicitly penalizes a deviation, teaching the ego to recover and steer back to the nominal path. A drifted state could in principle make this logged target infeasible to rejoin, which would manifest as either abrupt corrective maneuvers or failure to converge. We observe neither: relative to open-loop training, our approach reduces both jerk violations (Sec.~\ref{cl_paper_subsec:results_comfort}) and off-road rate (Sec.~\ref{cl_paper_subsec:results_main}), indicating that corrections remain smooth and effective.

%% file: sec/4_experiments.tex
\section{Experimental Setup}
\label{cl_paper_sec:experimental_setup}

We evaluate on nuScenes~\cite{caesar2020nuscenes}, \ac{av2}~\cite{wilson2023argoverse}, and DeepScenario~\cite{lu2023deepscenario}. The first two are standard trajectory forecasting benchmarks, while DeepScenario provides drone-recorded dense intersection scenarios that stress-test closed-loop training under frequent, close-proximity interactions. We pair these datasets with two model architectures: LMFormer-D, a decoder-only variant of LMFormer~\cite{yadav2025lmformer} described in more detail in Sec.~\ref{cl_paper_subsec:results_lmformer_d}, and an unmodified QCNet~\cite{zhou2023query}. The purpose of using two architectures is to confirm that the observed effects belong to the training paradigm rather than to a specific single model. In addition to standard open-loop prediction metrics, namely minFDE, minADE, and miss rate (MR) \cite{caesar2020nuscenes, wilson2023argoverse}, we evaluate closed-loop safety metrics, including collision rate, comfort, and off-road driving rate \cite{nuplan2021Caesar, gulino2023waymax}.

Unless stated otherwise, CL denotes our proposed training configuration: non-differentiable, on-policy closed-loop training with log-replay for surrounding agents. OL denotes an identical architecture and schedule trained with standard single-shot open-loop supervision. Both OL- and CL-trained models are evaluated in a closed-loop setting. To assess resilience against error accumulation, we sweep the evaluation replanning period $T_{\text{sim}} \in \{6.0, 2.0, 0.5\}\,\si{\second}$ for our main comparisons, testing whether CL training mitigates compounding errors at higher frequencies. Additionally, Sec.~\ref{cl_paper_subsec:ablation_diff} evaluates a finer sweep of $T_{sim}$ to characterize frequency coupling in greater detail. We report mean${}_{\pm\sigma}$ values over five random seeds for main results and three seeds for ablations.

%% file: sec/5_results.tex
\section{Results and Discussion}
\label{cl_paper_sec:results}

\subsection{Closed-Loop is better at higher replanning rates}
\label{cl_paper_subsec:results_main}

Tables~\ref{cl_paper_tab:nuscenes_main_results}--\ref{cl_paper_tab:qcnet_main_results} compare OL and CL training on three benchmarks that differ in architecture and datasets. Two use our LMFormer-D model (nuScenes, DeepScenario); the third applies our closed-loop training to an unmodified QCNet~\cite{zhou2023query} (\ac{av2}) to verify that the effect is a property of the training paradigm rather than the model.

\begin{table}[htbp]
\centering
\caption{OL vs.\ CL training with LMFormer-D on \textbf{nuScenes}. The CL advantage is negligible under single-shot evaluation and grows monotonically with the replanning rate.}
\label{cl_paper_tab:nuscenes_main_results}
\footnotesize
\setlength{\tabcolsep}{3.0pt}
\resizebox{\columnwidth}{!}{
\begin{tabular}{@{} c | c | *{2}{c} | *{2}{c} | c @{}}
\hline \noalign{\vspace{0.25mm}}
Eval & \multirow{2}{*}{Model} & \multicolumn{2}{c |}{k=1} & \multicolumn{2}{c |}{k=5} & \multirow{2}{*}{OffRoad\%$\downarrow$} \\
\cline{3-6} \noalign{\vspace{0.25mm}}
$T_{\text{sim}}$ &  & Col\%$\downarrow$ & minFDE$\downarrow$ & minADE$\downarrow$ & MR$\downarrow$ &  \\
\noalign{\vspace{0.25mm}} \hline \hline \noalign{\vspace{0.25mm}}
\multirow{2}{*}{6.0(s)} & OL & $2.97_{\pm.08}$ & $\mathbf{7.17}_{\pm.12}$ & $\mathbf{1.14}_{\pm.01}$ & $\mathbf{0.50}_{\pm.01}$ & $1.02_{\pm.05}$ \\
 & CL & $\mathbf{2.96}_{\pm.10}$ & $\mathbf{7.17}_{\pm.14}$ & $1.15_{\pm.02}$ & $0.51_{\pm.01}$ & $\mathbf{0.95}_{\pm.04}$ \\
\noalign{\vspace{0.25mm}} \hline \noalign{\vspace{0.25mm}}
\multirow{2}{*}{2.0(s)} & OL & $2.49_{\pm.14}$ & $7.18_{\pm.14}$ & $1.13_{\pm.01}$ & $0.50_{\pm.01}$ & $1.13_{\pm.12}$ \\
 & CL & $\mathbf{2.38}_{\pm.10}$ & $\mathbf{6.96}_{\pm.13}$ & $\mathbf{1.11}_{\pm.02}$ & $\mathbf{0.47}_{\pm.01}$ & $\mathbf{0.86}_{\pm.07}$ \\
\noalign{\vspace{0.25mm}} \hline \noalign{\vspace{0.25mm}}
\multirow{2}{*}{0.5(s)} & OL & $3.95_{\pm.19}$ & $8.37_{\pm.21}$ & $1.56_{\pm.03}$ & $0.66_{\pm.02}$ & $5.65_{\pm.69}$ \\
 & CL & $\mathbf{3.08}_{\pm.22}$ & $\mathbf{7.88}_{\pm.21}$ & $\mathbf{1.39}_{\pm.03}$ & $\mathbf{0.58}_{\pm.01}$ & $\mathbf{1.86}_{\pm.60}$ \\
\noalign{\vspace{0.25mm}} \hline \hline
\end{tabular}
}
\vspace{-1.5mm}
\end{table}

\begin{table}[htbp]
\centering
\caption{OL vs.\ CL training on \textbf{DeepScenario}. Denser, more interactive scenarios amplify the gains of CL at higher replanning rates.}
\label{cl_paper_tab:deepScenario_main_results}
\footnotesize
\setlength{\tabcolsep}{2.8pt}
\resizebox{\columnwidth}{!}{
\begin{tabular}{@{} c | c | *{2}{c} | *{2}{c} | c @{}}
\hline \noalign{\vspace{0.25mm}}
Eval & \multirow{2}{*}{Model} & \multicolumn{2}{c |}{k=1} & \multicolumn{2}{c |}{k=5} & \multirow{2}{*}{OffRoad\%$\downarrow$} \\
\cline{3-6} \noalign{\vspace{0.25mm}}
$T_{\text{sim}}$ &  & Col\%$\downarrow$ & minFDE$\downarrow$ & minADE$\downarrow$ & MR$\downarrow$ &  \\
\noalign{\vspace{0.25mm}} \hline \hline \noalign{\vspace{0.25mm}}
\multirow{2}{*}{6.0(s)} & OL & $2.23_{\pm.30}$ & $7.50_{\pm.34}$ & $\mathbf{1.10}_{\pm.05}$ & $\mathbf{0.45}_{\pm.04}$ & $7.26_{\pm.88}$ \\
 & CL & $\mathbf{1.90}_{\pm.28}$ & $\mathbf{6.84}_{\pm.23}$ & $1.19_{\pm.07}$ & $0.49_{\pm.02}$ & $\mathbf{6.03}_{\pm.90}$ \\
\noalign{\vspace{0.25mm}} \hline \noalign{\vspace{0.25mm}}
\multirow{2}{*}{2.0(s)} & OL & $1.75_{\pm.05}$ & $7.63_{\pm.28}$ & $\mathbf{1.24}_{\pm.06}$ & $0.48_{\pm.02}$ & $5.64_{\pm.58}$ \\
 & CL & $\mathbf{1.33}_{\pm.24}$ & $\mathbf{6.99}_{\pm.31}$ & $\mathbf{1.24}_{\pm.14}$ & $\mathbf{0.44}_{\pm.05}$ & $\mathbf{4.72}_{\pm1.45}$ \\
\noalign{\vspace{0.25mm}} \hline \noalign{\vspace{0.25mm}}
\multirow{2}{*}{0.5(s)} & OL & $4.46_{\pm.20}$ & $12.61_{\pm.83}$ & $3.22_{\pm.07}$ & $0.90_{\pm.02}$ & $22.28_{\pm4.04}$ \\
 & CL & $\mathbf{2.92}_{\pm.42}$ & $\mathbf{10.93}_{\pm1.08}$ & $\mathbf{3.01}_{\pm.53}$ & $\mathbf{0.81}_{\pm.04}$ & $\mathbf{8.62}_{\pm2.37}$ \\
\noalign{\vspace{0.25mm}} \hline \hline
\end{tabular}
}
\end{table}

\begin{table}[htbp]
\centering
\caption{OL vs.\ CL training with \textbf{QCNet} on
\textbf{Argoverse~2}. CL improves mean of reported metrics at $T_{\text{sim}}=\SI{0.5}{\second}$}
\label{cl_paper_tab:qcnet_main_results}
\footnotesize
\setlength{\tabcolsep}{3.1pt}
\resizebox{\columnwidth}{!}{
\begin{tabular}{@{}  c | c | c | c | c | c | c @{}}
\hline \noalign{\vspace{0.25mm}}
Eval & \multirow{2}{*}{Model} & \multicolumn{2}{c |}{k=1} & \multicolumn{2}{c |}{k=6} & \multirow{2}{*}{OffRoad\%$\downarrow$} \\
\cline{3-6} \noalign{\vspace{0.25mm}}
$T_{\text{sim}}$ &  & Col\%$\downarrow$ & minFDE$\downarrow$ & minADE$\downarrow$ & MR$\downarrow$ &  \\
\noalign{\vspace{0.25mm}} \hline \hline \noalign{\vspace{0.25mm}}
\multirow{2}{*}{6.0(s)} & OL & $\mathbf{1.64}_{\pm.09}$ & $\mathbf{4.44}_{\pm.04}$ & $0.76_{\pm.01}$ & $\mathbf{0.18}_{\pm.00}$ & $\mathbf{11.76}_{\pm.29}$ \\
 & CL & $1.69_{\pm.06}$ & $4.75_{\pm.05}$ & $\mathbf{0.75}_{\pm.00}$ & $0.19_{\pm.00}$ & $11.99_{\pm.12}$ \\
\noalign{\vspace{0.25mm}} \hline \noalign{\vspace{0.25mm}}
\multirow{2}{*}{2.0(s)} & OL & $1.98_{\pm.04}$ & $5.47_{\pm.25}$ & $0.93_{\pm.03}$ & $0.29_{\pm.03}$ & $13.66_{\pm.50}$ \\
 & CL & $\mathbf{1.87}_{\pm.06}$ & $\mathbf{5.40}_{\pm.13}$ & $\mathbf{0.89}_{\pm.01}$ & $\mathbf{0.28}_{\pm.00}$ & $\mathbf{13.13}_{\pm.52}$ \\
\noalign{\vspace{0.25mm}} \hline \noalign{\vspace{0.25mm}}
\multirow{2}{*}{0.5(s)} & OL & $2.84_{\pm.14}$ & $8.35_{\pm.47}$ & $1.30_{\pm.01}$ & $0.52_{\pm.03}$ & $16.67_{\pm.67}$ \\
 & CL & $\mathbf{2.35}_{\pm.31}$ & $\mathbf{7.28}_{\pm.46}$ & $\mathbf{1.14}_{\pm.05}$ & $\mathbf{0.44}_{\pm.03}$ & $\mathbf{15.07}_{\pm.52}$ \\
\noalign{\vspace{0.25mm}} \hline \hline
\end{tabular}
}
\vspace{-2mm}
\end{table}

At $T_{\text{sim}}=\SI{6.0}{\second}$, OL and CL are comparable across all three benchmarks, confirming that closed-loop training does not trade away single-shot accuracy. The CL margin then grows monotonically as $T_{\text{sim}}$ decreases: at maximum replanning rate, collisions fall by $22\%$ (nuScenes), $35\%$ (DeepScenario), and $17\%$ (\ac{av2}), and off-road rate by $67\%$, $61\%$, and $10\%$ respectively.

Two properties of the \ac{av2} setting merit comment, as the performance gains from \ac{cl} training are least pronounced here. It records motion at 10 Hz compared to nuScenes’ or DeepScenario's 2 Hz, creating five times more trajectory points. We hypothesize that, this finer resolution tightly constrains predictions, preventing OL models from drifting (minFDE$_\mathrm{1}$ in single shot-evaluation is 4.44 vs. 7.17 for nuScenes and 7.50 for DeepScenario). Consequently, CL models only show a clear advantage at highest replanning rate where errors compound rapidly. Additionally, because \ac{av2} lacks exact bounding-box data for individual vehicles, safety checks (collision and offroad) rely on standard class averages, adding equal baseline noise to both OL and CL.

\subsection{Where the Error Accumulates}
\label{cl_paper_subsec:results_rollout}

\begin{figure}[htbp]
    \centering
    \includegraphics[width=\linewidth]{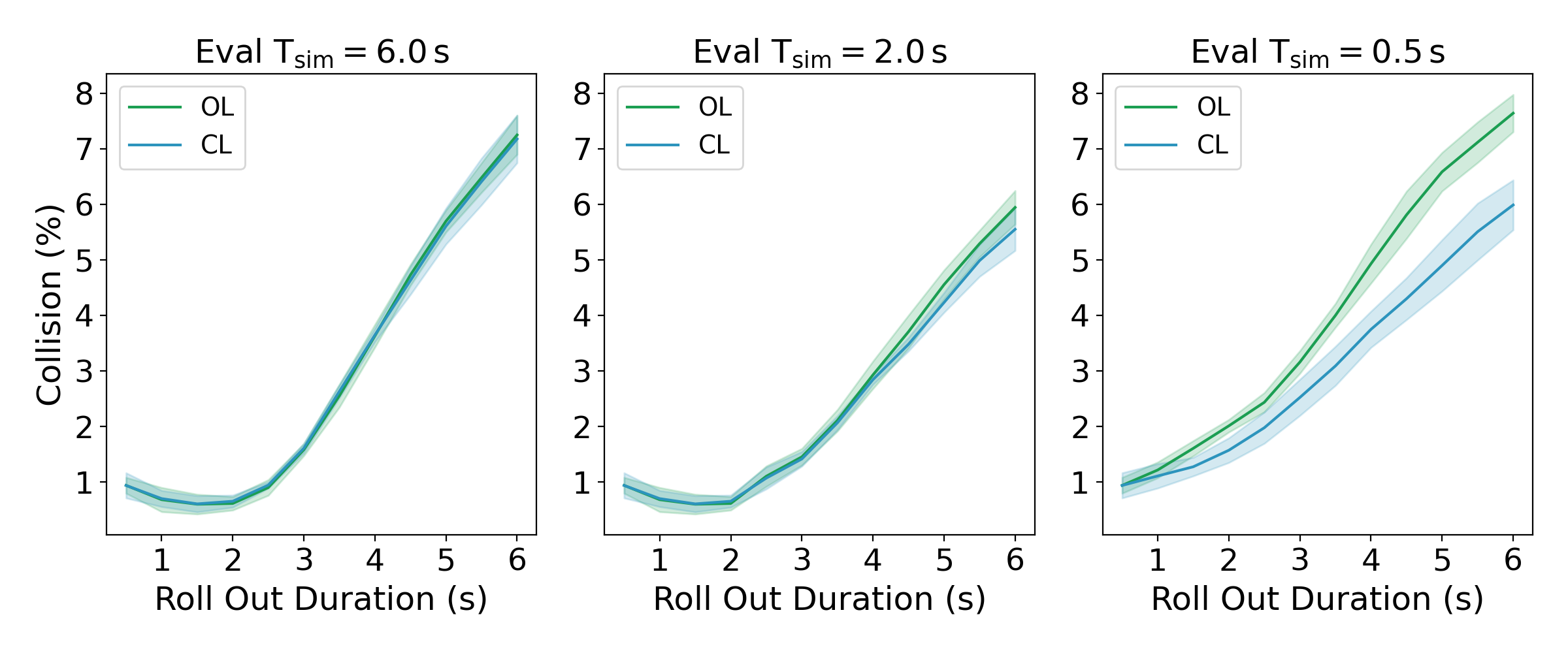}
    \caption{Collision rate vs.\ rollout duration on nuScenes. Under single-shot evaluation (left) the OL and CL curves coincide. At $T_{\text{sim}}=\SI{0.5}{\second}$ (right) the curves are indistinguishable for the first \SI{1.5}{\second} and then diverge, reaching a $1.7$\,pp gap at \SI{6}{\second}. The gap is therefore not a constant improvement in prediction quality but a difference in the \emph{rate} at which error compounds.}
    \label{cl_paper_fig:col_rollout}
\end{figure}

Previously reported aggregate metrics over complete trajectories do not highlight when the two regimes begin to diverge temporally. To capture these temporal dynamics, Figure~\ref{cl_paper_fig:col_rollout} plots collision rates over elapsed rollout time.

Under single-shot evaluation, the OL and CL curves coincide, as expected. At $T_{\text{sim}}=\SI{0.5}{\second}$, both models initially exhibit identical collision rates. However, their performance diverges as the rollout progresses. This widening gap over time demonstrates that the primary benefit of CL training lies in mitigating error accumulation in long-tail scenarios under distribution shift, rather than improving single-shot accuracy.

\subsection{Comfort Analysis}
\label{cl_paper_subsec:results_comfort}

\begin{figure}[htbp]
    \centering
    \includegraphics[width=\linewidth]{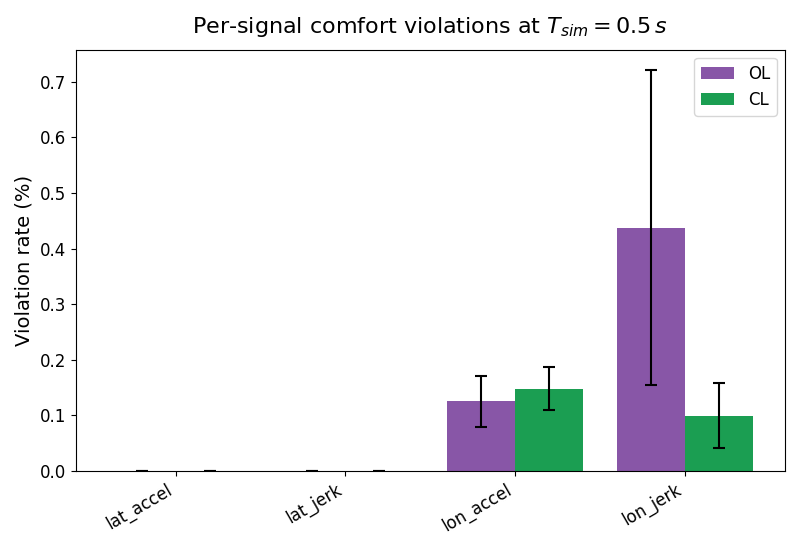}
    \caption{Per-signal comfort violation rate at $T_{\text{sim}}=\SI{0.5}{\second}$ averaged over all five modes and twelve timesteps. Lateral violations are negligible for both. Longitudinal acceleration is comparable, however, longitudinal jerk is substantially lower for CL, indicating no corrective jerk from supervising against the original logged trajectory. Error bars: $\pm 1\sigma$.}
    \label{cl_paper_fig:comfort_violation_tsim_0.5}
\end{figure}

A natural concern with our formulation is that supervision against the \emph{original} recorded future from a drifted state may induce abrupt corrective swings that would register as comfort violations. To test this, we evaluate comfort violations at $T_{\text{sim}}=\SI{0.5}{\second}$ on nuScenes (Fig.~\ref{cl_paper_fig:comfort_violation_tsim_0.5}), where corrective updates are most frequent. 

Although lateral violations are negligible for both configurations, this does not imply that lateral error is absent. In fact, the off-road rate improves with the CL configuration (Table~\ref{cl_paper_tab:nuscenes_main_results}). Comfort thresholds respond to the \emph{rate} of correction rather than the accumulated offset, so slow lateral drift can integrate into a lane departure without ever triggering a comfort violation. The primary signal is only longitudinal: CL violates the acceleration threshold marginally more ($0.15_{\pm.04}\%$ vs.\ $0.125_{\pm.05}\%$) but violates the jerk threshold substantially less ($0.10_{\pm.06}\%$ vs.\ $0.44_{\pm.28}\%$). CL is thus the smoother policy on the metric most sensitive to abrupt corrections, i.e., jerk. We attribute this to the reduced drift of the CL model (Sec.~\ref{cl_paper_subsec:results_rollout}), which keeps necessary corrections small and gradual. Conversely, OL, lacking recovery behavior, drifts to larger offsets and produces sharper corrections that register as jerk.

\subsection{Ablation: Off-Policy vs.\ On-Policy Rollout}
\label{cl_paper_subsec:ablation_on_off_policy}

We isolate the role of the rollout distribution by comparing the on-policy configuration against an off-policy variant in which the best mode is re-selected at each replanning step during training, forcing one mode to learn from another's mistakes (Table~\ref{cl_paper_tab:off_vs_on_policy_nus}).

\begin{table}[htbp]
\centering
\caption{\textbf{Off- vs.\ on-policy} closed-loop rollout during training (nuScenes, three seeds). On-policy rollout dominates every displacement and compliance metric at $T_{\text{sim}} \leq \SI{2.0}{\second}$.}
\label{cl_paper_tab:off_vs_on_policy_nus}
\footnotesize
\setlength{\tabcolsep}{3.0pt}
\resizebox{\columnwidth}{!}{
\begin{tabular}{@{} c | c | *{2}{c} | *{2}{c} | c @{}}
\hline \noalign{\vspace{0.25mm}}
Eval & \multirow{2}{*}{Policy} & \multicolumn{2}{c |}{k=1} & \multicolumn{2}{c |}{k=5} & \multirow{2}{*}{OffRoad\%$\downarrow$} \\
\cline{3-6} \noalign{\vspace{0.25mm}}
$T_{\text{sim}}$ &  & Col\%$\downarrow$ & minFDE$\downarrow$ & minADE$\downarrow$ & MR$\downarrow$ &  \\
\noalign{\vspace{0.25mm}} \hline \hline \noalign{\vspace{0.25mm}}
\multirow{2}{*}{6.0(s)} & Off- & $\mathbf{2.83}_{\pm.20}$ & $\mathbf{7.02}_{\pm.12}$ & $\mathbf{1.15}_{\pm.00}$ & $\mathbf{0.50}_{\pm.01}$ & $1.06_{\pm.04}$ \\
 & On- & $2.96_{\pm.10}$ & $7.17_{\pm.14}$ & $\mathbf{1.15}_{\pm.02}$ & $0.51_{\pm.01}$ & $\mathbf{0.95}_{\pm.04}$ \\
\noalign{\vspace{0.25mm}} \hline \noalign{\vspace{0.25mm}}
\multirow{2}{*}{2.0(s)} & Off- & $\mathbf{1.92}_{\pm.04}$ & $7.60_{\pm.01}$ & $1.28_{\pm.02}$ & $0.58_{\pm.02}$ & $1.42_{\pm.02}$ \\
 & On- & $2.38_{\pm.10}$ & $\mathbf{6.96}_{\pm.13}$ & $\mathbf{1.11}_{\pm.02}$ & $\mathbf{0.47}_{\pm.01}$ & $\mathbf{0.86}_{\pm.07}$ \\
\noalign{\vspace{0.25mm}} \hline \noalign{\vspace{0.25mm}}
\multirow{2}{*}{0.5(s)} & Off- & $3.50_{\pm.05}$ & $12.51_{\pm.83}$ & $2.35_{\pm.11}$ & $0.77_{\pm.02}$ & $2.47_{\pm1.07}$ \\
 & On- & $\mathbf{3.08}_{\pm.22}$ & $\mathbf{7.88}_{\pm.21}$ & $\mathbf{1.39}_{\pm.03}$ & $\mathbf{0.58}_{\pm.01}$ & $\mathbf{1.86}_{\pm.60}$ \\
\noalign{\vspace{0.25mm}} \hline \hline
\end{tabular}
}
\end{table}

At $T_{\text{sim}}=\SI{0.5}{\second}$, off-policy training yields significantly higher displacement errors, increasing minFDE by $59\%$ and minADE by $69\%$, while collision rates rise by only $14\%$. This asymmetry suggests that cross-mode supervision degrades trajectory accuracy far more than it impacts collision risk. By supervising each mode strictly with its own rollout states, on-policy training eliminates off-target signals, allowing the model to better generalize to high replanning rates.

\subsection{Ablation: Differentiable vs.\ Non-Differentiable Rollout}
\label{cl_paper_subsec:ablation_diff}

\begin{figure*}[htbp]
    \centering
    \begin{subfigure}[b]{0.32\textwidth}
        \centering
        \includegraphics[width=\textwidth]{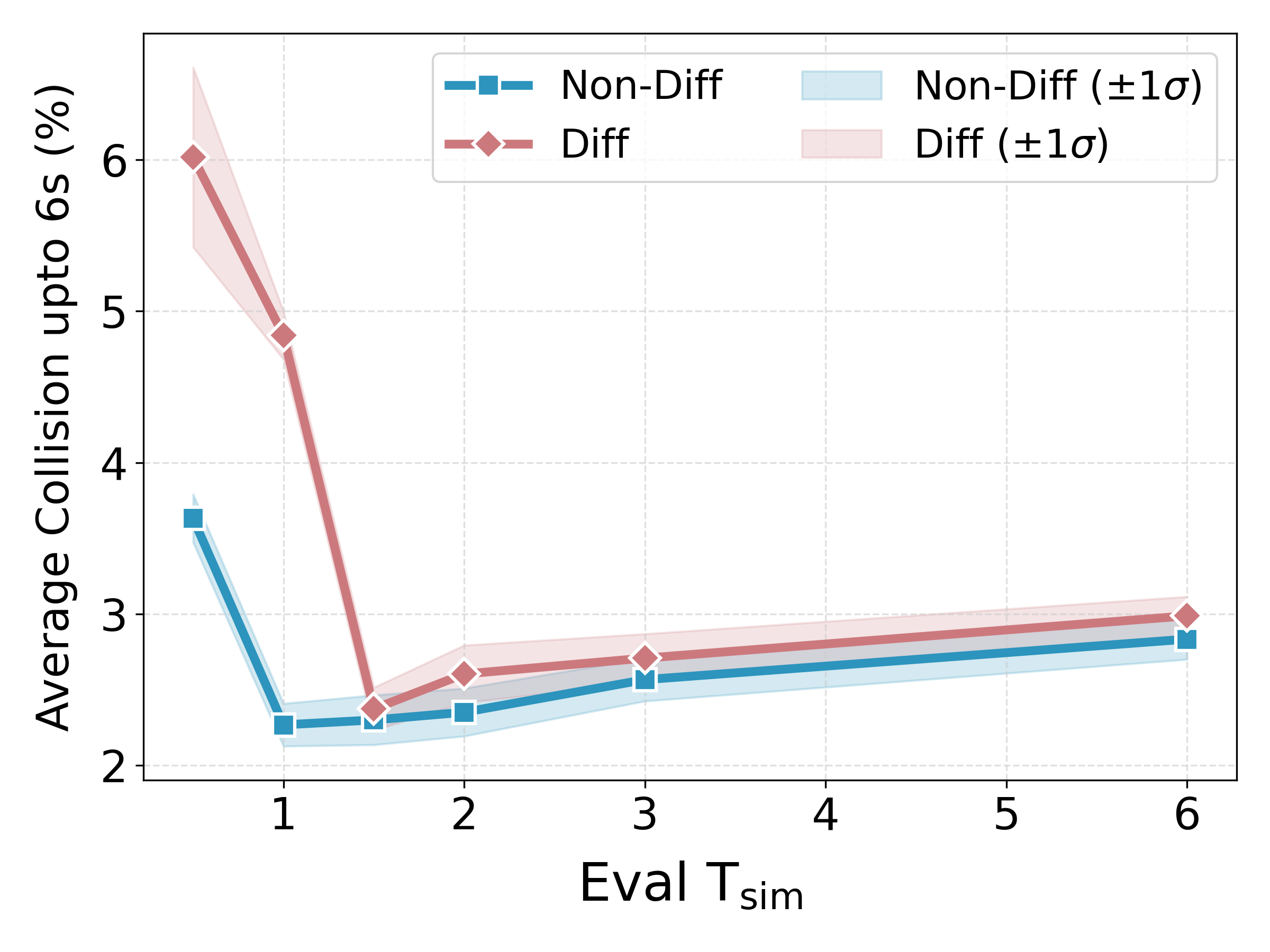}
        \caption{Train $T_{\text{sim}} = \SI{1.5}{\second}$}
        \label{cl_paper_fig:sub1}
    \end{subfigure}
    \hfill
    \begin{subfigure}[b]{0.32\textwidth}
        \centering
        \includegraphics[width=\textwidth]{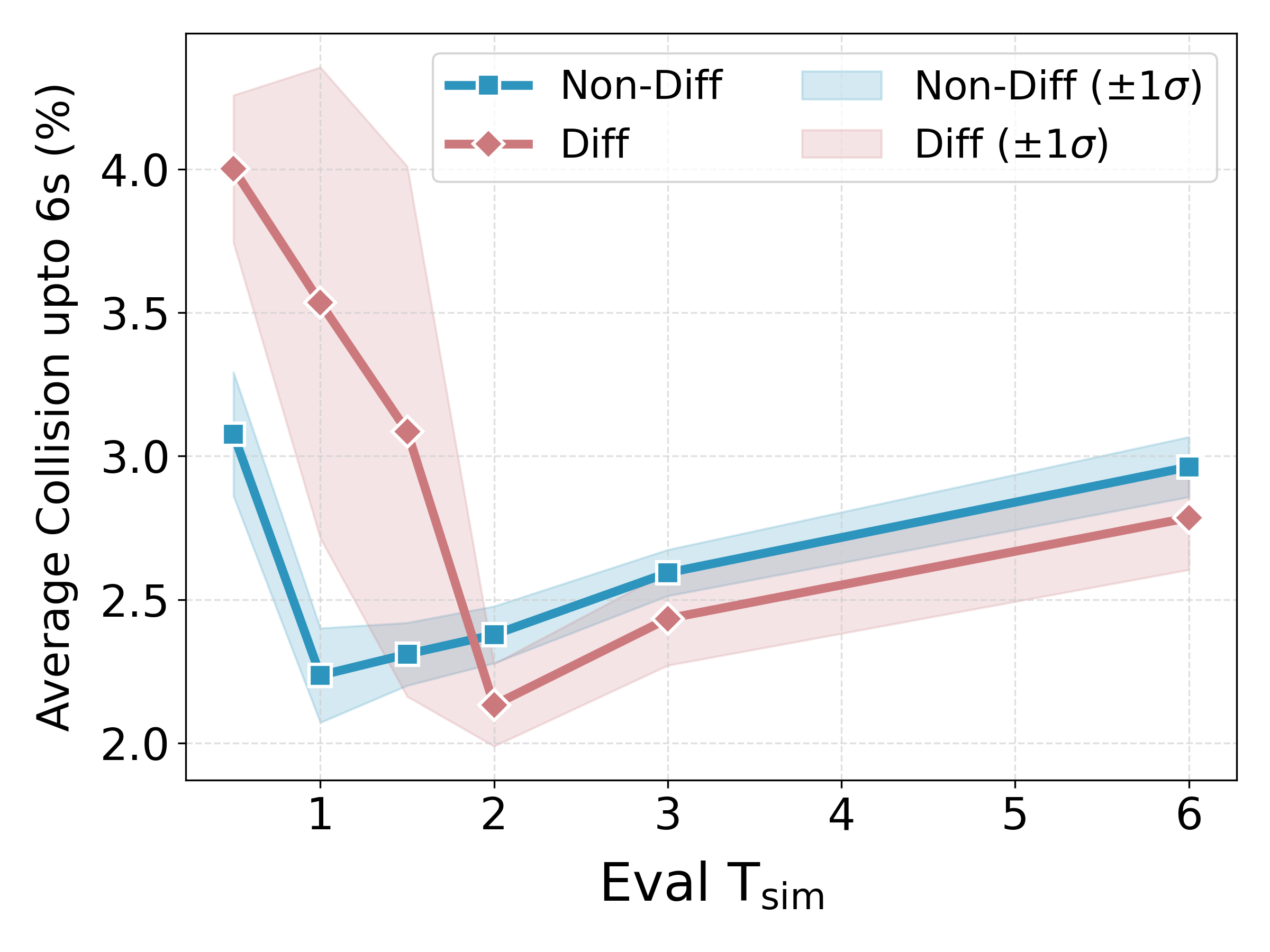}
        \caption{Train $T_{\text{sim}} = \SI{2.0}{\second}$}
        \label{cl_paper_fig:sub2}
    \end{subfigure}
    \hfill
    \begin{subfigure}[b]{0.32\textwidth}
        \centering
        \includegraphics[width=\textwidth]{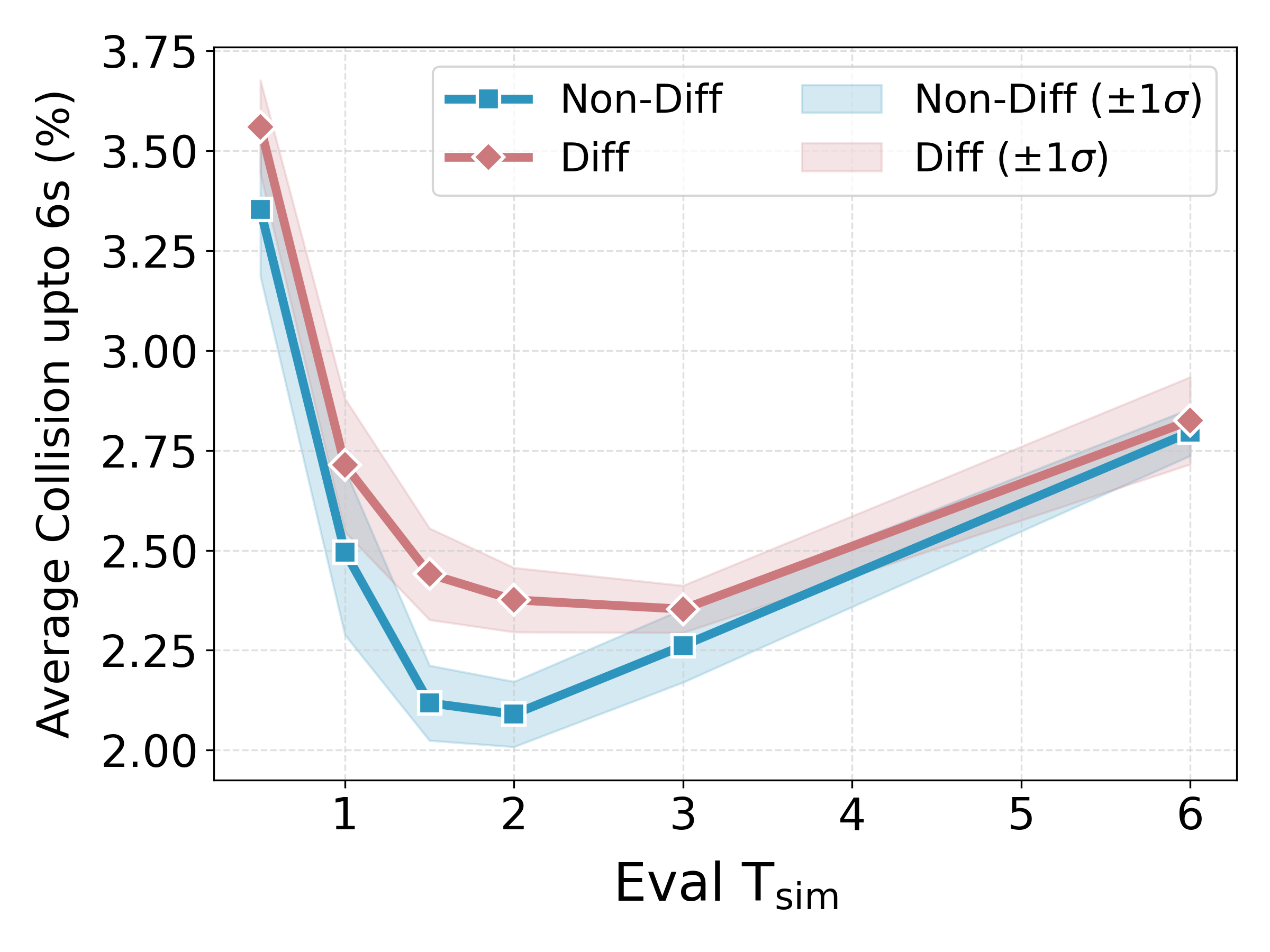}
        \caption{Train $T_{\text{sim}} = \SI{3.0}{\second}$}
        \label{cl_paper_fig:sub3}
    \end{subfigure}
    \caption{\textbf{Differentiable training produces a frequency-specific policy.} Collision rate over a \SI{6}{\second} rollout versus evaluation replanning period for CL models trained at three different replanning periods. The \emph{Diff} model's minimum collision rate (red) tracks its training $T_{\text{sim}}$ in every panel. In contrast, \emph{Non-Diff} models (blue) generalize better to higher replanning rates, with the $T_{\text{sim}} = \SI{2.0}{\second}$ variant achieving the lowest overall collision rates.}
    \label{cl_paper_fig:diff_vs_nondiff}
    \vspace{-2mm}
\end{figure*}

We test the effect of differentiable training by removing the \texttt{stop\_gradient} operator from Alg.~\ref{cl_paper_alg:method_cl_training}. Fig.~\ref{cl_paper_fig:diff_vs_nondiff} compares the resulting \emph{Diff} variant against our proposed \emph{Non-Diff} configuration for three CL models trained at $T_{\text{sim}} \in \{1.5, 2.0, 3.0\}\,\si{\second}$, each evaluated across a sweep of replanning periods from $\SI{0.5}{\second}$ to $\SI{6.0}{\second}$.

The pattern is consistent across all three CL models: the Diff collision minimum tracks its training replanning period exactly, while the Non-Diff minimum generalizes to higher replanning rates regardless of training configuration. This is consistent with the gradient shortcut hypothesized in Sec.~\ref{cl_paper_sec:introduction}: backpropagation through the executed transition may let the model retroactively adjust an earlier prediction to make the resulting state easier to predict from, rather than genuinely correcting for drift, coupling it to its training frequency. Detaching this transition (Non-Diff) severs this shortcut, forcing each cycle to handle real drift independently and allowing the resulting policy to generalize across replanning rates rather than locking onto the one used in training. This reduced frequency coupling is particularly valuable for deployment, as real-world systems benefit from running at higher replanning rates for improved safety and responsiveness. However, training directly at high rates requires additional $N$ forward passes for $N$ closed-loop samples, making it increasingly expensive. Generalization across rates offers a way around this: a model trained at a low, cheaper rate can be deployed at higher rates without retraining.

The Non-Diff curves reveal a second effect: while the Non-Diff model generalizes to higher evaluation replanning rates, its absolute performance still depends on the training replanning period. Across all three panels, the model trained at $T_{\text{sim}}=\SI{2.0}{\second}$ achieves the lowest collisions at high replanning rates ($T_{\text{sim}} \leq \SI{1.0}{\second}$), outperforming both the $\SI{1.5}{\second}$ and $\SI{3.0}{\second}$ variants. We attribute this to a fundamental trade-off in choosing the training replanning period. This interval must be long enough to expose the model to compounding errors, yet short enough to leave a sufficient remaining supervision horizon ($T_\text{pred} - n  T_\text{sim}$) for closed-loop samples, which is why $T_{sim} = \SI{2.0}{\second}$ is adopted as the default closed-loop training configuration.

\subsection{Ablation: Log-Replay vs.\ Scene Prediction}
\label{cl_paper_subsec:ablation_log_vs_scene}

During CL training, surrounding agent rollouts are executed via log-replay. A natural extension is to simulate surrounding agents using the prediction model itself, with the aim of making them reactive to the ego vehicle. To maintain scene consistency (ensuring surrounding agents' predicted trajectories remain plausible with respect to their actual eventual positions rather than drifting into arbitrary modes), we condition their predictions on their final recorded goal positions, yielding unimodal trajectories. This attempt did not succeed: the goal-conditioning dominates the learning signal, and rather than reacting to the ego, surrounding agents closely replicate their log-replay trajectories. Accordingly, Table~\ref{cl_paper_tab:log_vs_fullyScene} shows parity: scene prediction is marginally ahead at $T_{\text{sim}} \geq \SI{2.0}{\second}$ and marginally behind at $\SI{0.5}{\second}$, with all differences within seed variance. However, this only compares two variants that are both effectively non-reactive, and does not establish that log-replay is equivalent to a genuinely reactive simulator. 

\begin{table}[htbp]
\centering
\caption{\textbf{Log-replay vs.\ goal-conditioned scene prediction} during training on nuScenes. Differences are within or close to the seed spread.}
\label{cl_paper_tab:log_vs_fullyScene}
\footnotesize
\setlength{\tabcolsep}{3.4pt}
\resizebox{\columnwidth}{!}{
\begin{tabular}{@{} c | c | *{2}{c} | *{2}{c} | c @{}}
\hline \noalign{\vspace{0.25mm}}
Eval & \multirow{2}{*}{Sim.} & \multicolumn{2}{c |}{k=1} & \multicolumn{2}{c |}{k=5} & \multirow{2}{*}{OffRoad\%$\downarrow$} \\
\cline{3-6} \noalign{\vspace{0.25mm}}
$T_{\text{sim}}$ &  & Col\%$\downarrow$ & minFDE$\downarrow$ & minADE$\downarrow$ & MR$\downarrow$ &  \\
\noalign{\vspace{0.25mm}} \hline \hline \noalign{\vspace{0.25mm}}
\multirow{2}{*}{6.0(s)} & Scene & $\mathbf{2.89}_{\pm.11}$ & $7.22_{\pm.21}$ & $\mathbf{1.15}_{\pm.01}$ & $\mathbf{0.51}_{\pm.01}$ & $\mathbf{0.94}_{\pm.07}$ \\
 & Log & $2.96_{\pm.10}$ & $\mathbf{7.17}_{\pm.14}$ & $\mathbf{1.15}_{\pm.02}$ & $\mathbf{0.51}_{\pm.01}$ & $0.95_{\pm.04}$ \\
\noalign{\vspace{0.25mm}} \hline \noalign{\vspace{0.25mm}}
\multirow{2}{*}{2.0(s)} & Scene & $\mathbf{2.24}_{\pm.10}$ & $\mathbf{6.93}_{\pm.24}$ & $\mathbf{1.11}_{\pm.01}$ & $\mathbf{0.47}_{\pm.01}$ & $\mathbf{0.78}_{\pm.08}$ \\
 & Log & $2.38_{\pm.10}$ & $6.96_{\pm.13}$ & $\mathbf{1.11}_{\pm.02}$ & $\mathbf{0.47}_{\pm.01}$ & $0.86_{\pm.07}$ \\
\noalign{\vspace{0.25mm}} \hline \noalign{\vspace{0.25mm}}
\multirow{2}{*}{0.5(s)} & Scene & $3.35_{\pm.44}$ & $\mathbf{7.85}_{\pm.12}$ & $1.43_{\pm.05}$ & $0.59_{\pm.02}$ & $2.14_{\pm.37}$ \\
 & Log & $\mathbf{3.08}_{\pm.22}$ & $7.88_{\pm.21}$ & $\mathbf{1.39}_{\pm.03}$ & $\mathbf{0.58}_{\pm.01}$ & $\mathbf{1.86}_{\pm.60}$ \\
\noalign{\vspace{0.25mm}} \hline \hline
\end{tabular}
}
\end{table}

\begin{table*}[h]
\vspace{2mm}
\centering
\footnotesize
\setlength{\tabcolsep}{3.4pt}
\caption{Effect of the decoder-only design on nuScenes. Prediction quality is preserved while parameters and FLOPs are roughly halved and thirded.}
\label{cl_paper_tab:decoder_only}
\small
\begin{tabular}{@{} c | *{3}{c} | *{3}{c} | c | *{2}{c} @{}}
\hline \noalign{\vspace{0.25mm}}
\multirow{3}{*}{Models} & \multicolumn{7}{c |}{Prediction Metrics} & \multicolumn{2}{c}{Model Efficiency} \\
\cline{2-10} \noalign{\vspace{0.5mm}}
 & \multicolumn{3}{c|}{k=5} & \multicolumn{3}{c |}{k=1} & OffRoad & Parameters & FLOPS \\
\cline{2-7} \noalign{\vspace{0.5mm}}
 & minFDE$\downarrow$ & minADE$\downarrow$ & MR$\downarrow$ & minFDE$\downarrow$ & minADE$\downarrow$ & MR$\downarrow$ & Rate$\downarrow$ & (mil) & (GFLOPS) \\
\hline \hline \noalign{\vspace{0.5mm}}
LMFormer\tnote{1} & 2.145 & 1.155 & 0.500 & 7.114 & \textbf{3.136} & 0.832 & \textbf{0.010} & 2.234 & 3.365 \\
LMFormer-D\tnote{1} & \textbf{2.141} & \textbf{1.141} & \textbf{0.492} & \textbf{7.069} & 3.152 & \textbf{0.829} & 0.011 & \textbf{1.207} (-45\%) & \textbf{1.249} (-62\%) \\
\hline \hline
\end{tabular}
\end{table*}

\begin{figure*}[h]
    \centering
    \setlength{\arrayrulewidth}{0.8pt}
    \setlength{\tabcolsep}{1pt}
    \begin{tabular}{|c|c|c|c|c|}
        \hline
        \begin{subfigure}[t]{0.19\linewidth} 
            \includegraphics[width=\linewidth,trim={250 150 200 335},clip]{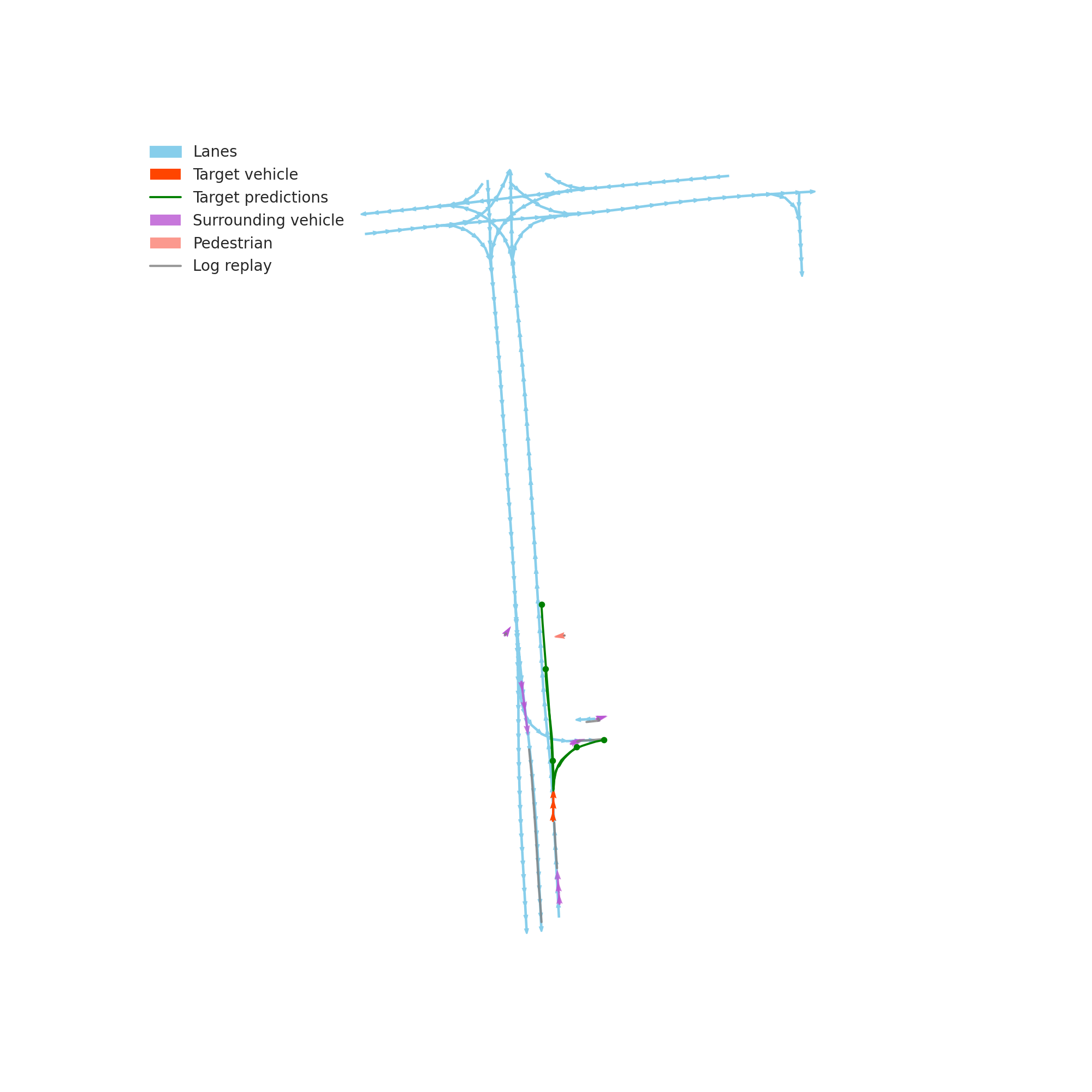}
        \end{subfigure}
        & 
        \begin{subfigure}[t]{0.19\linewidth} 
            \includegraphics[width=\linewidth,trim={250 150 200 335},clip]{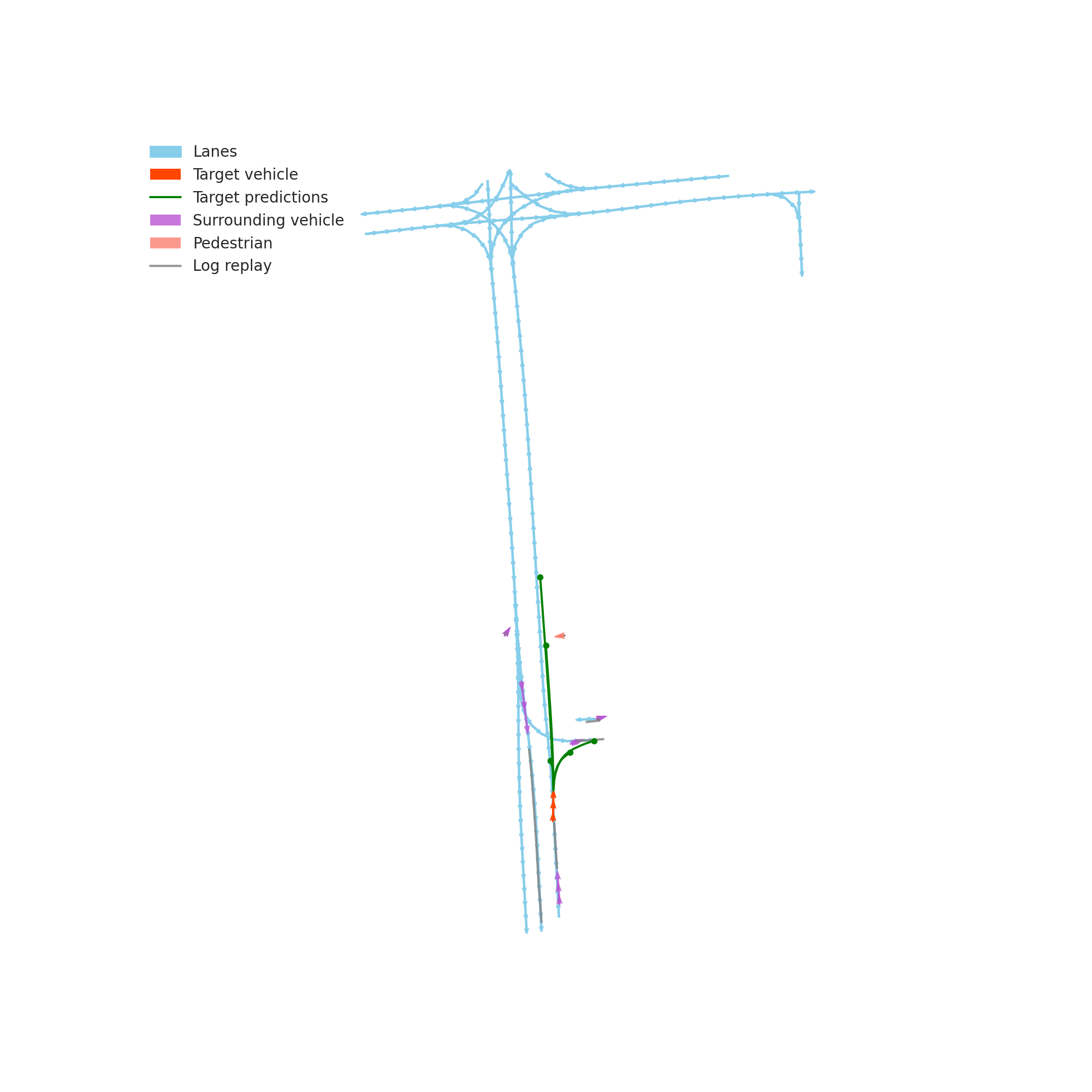}
        \end{subfigure}
        &
        \begin{subfigure}[t]{0.19\linewidth} 
            \includegraphics[width=\linewidth,trim={250 150 200 335},clip]{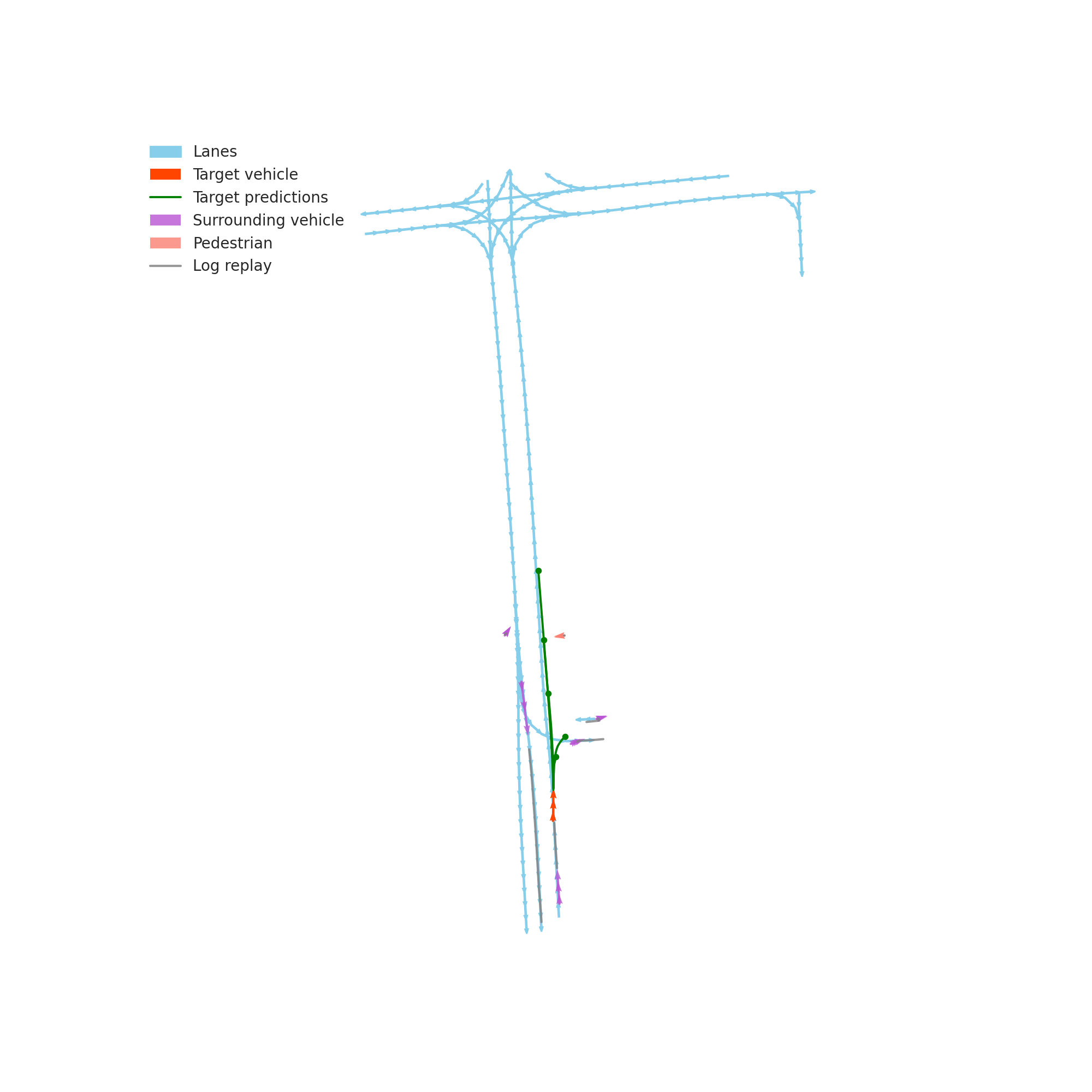}
        \end{subfigure}
        &
        \begin{subfigure}[t]{0.19\linewidth} 
            \includegraphics[width=\linewidth,trim={250 150 200 335},clip]{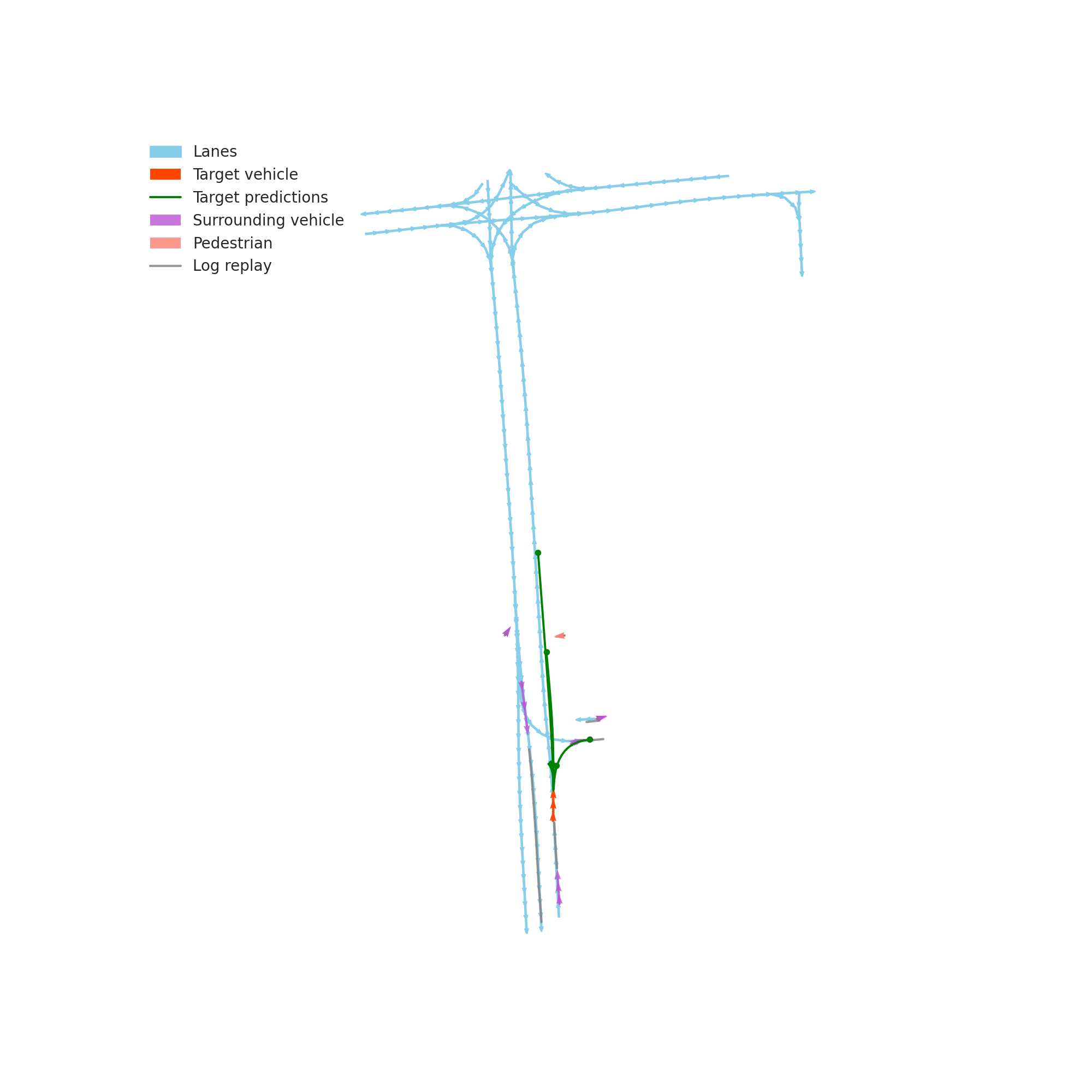}
        \end{subfigure}
        &
        \begin{subfigure}[t]{0.19\linewidth} 
            \begin{tikzpicture}
                \node[inner sep=0pt] (mainpic) {
                       \includegraphics[width=\linewidth,trim={300 150 150 335},clip]{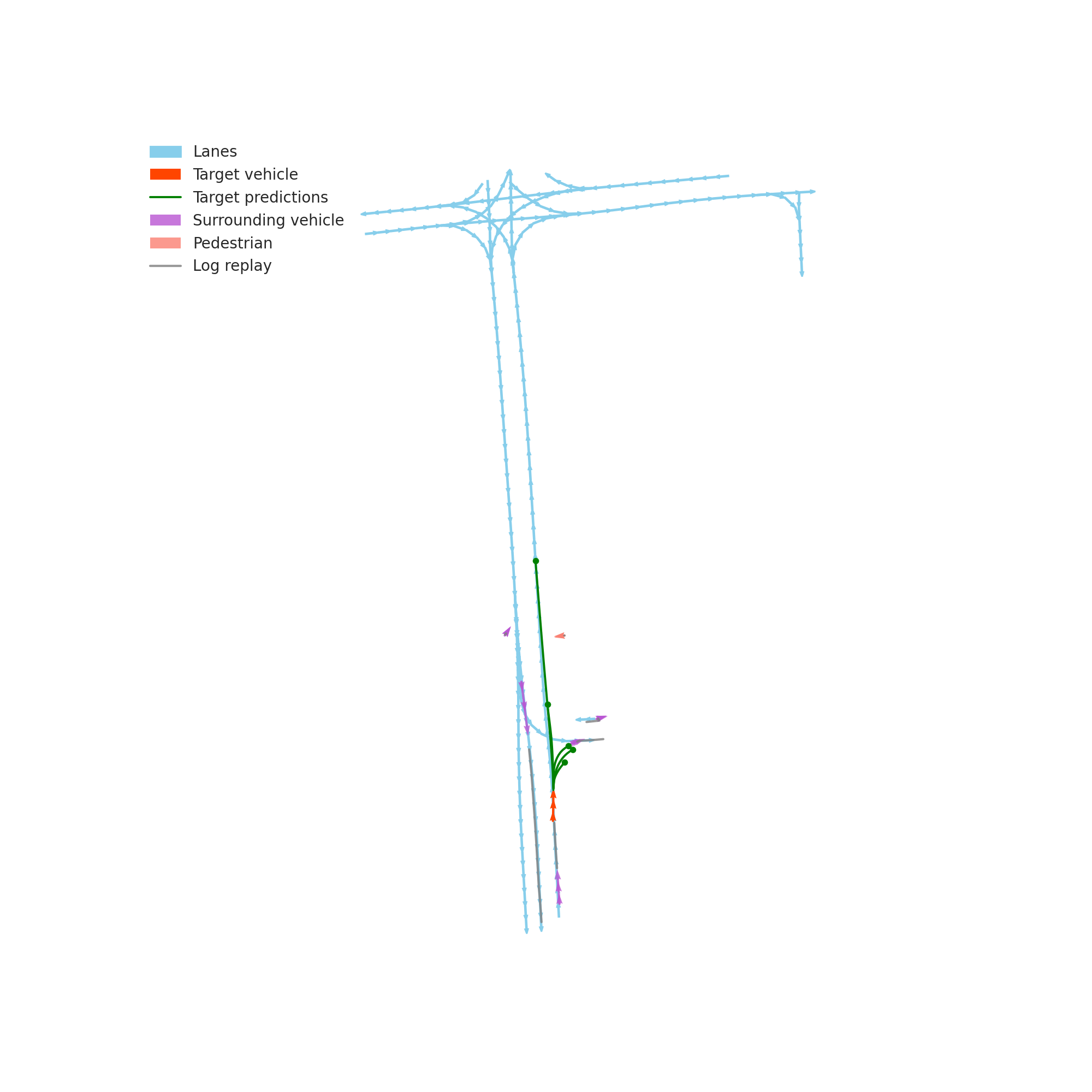}
                };
                   \node[anchor=north east, at=(mainpic.north east)] {
                       \includegraphics[width=0.6\linewidth]{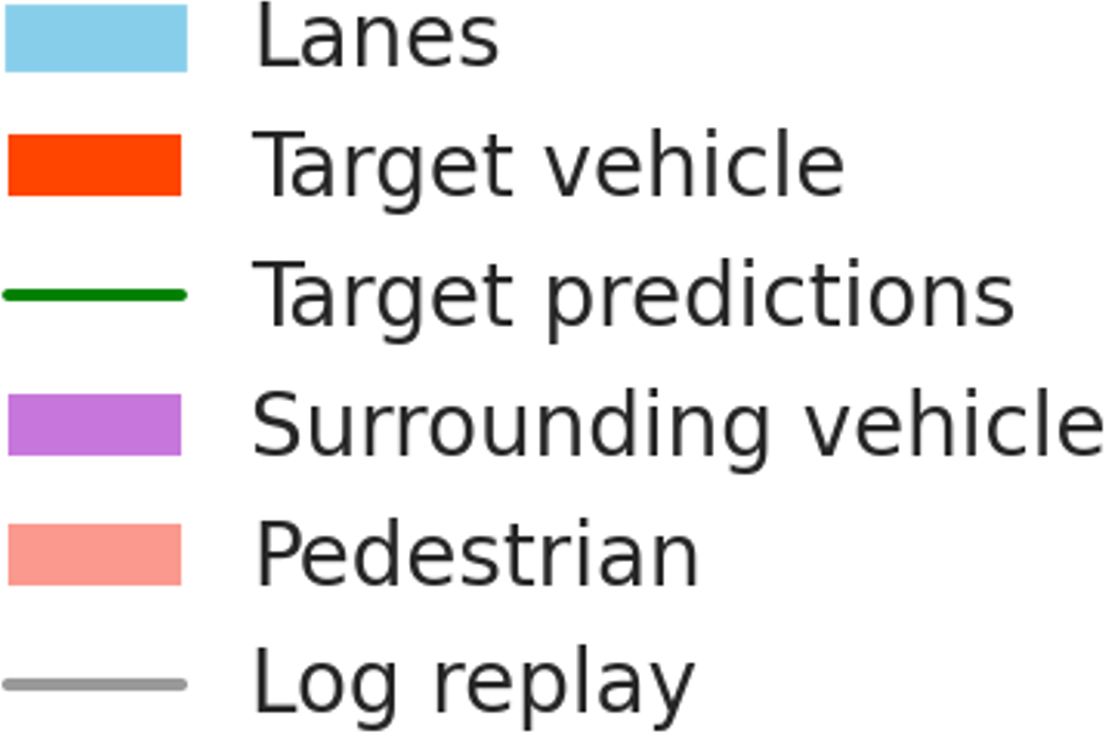}
                };
            \end{tikzpicture}
        \end{subfigure}
        \\[-4pt]
        \hline 
        \begin{subfigure}[t]{0.19\linewidth} 
            \includegraphics[width=\linewidth,trim={250 145 200 325},clip]{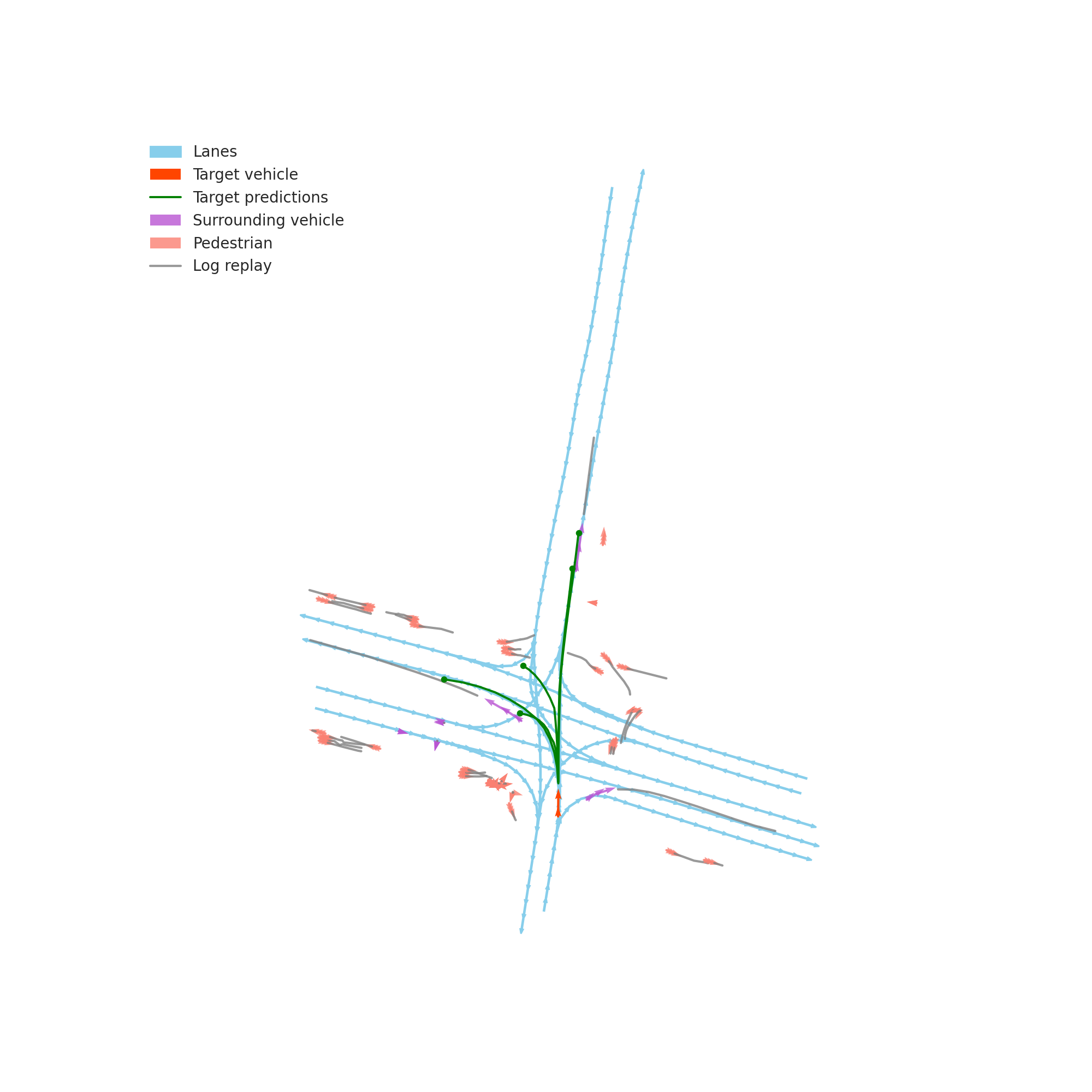}
        \end{subfigure}
        &
        \begin{subfigure}[t]{0.19\linewidth} 
            \includegraphics[width=\linewidth,trim={250 145 200 325},clip]{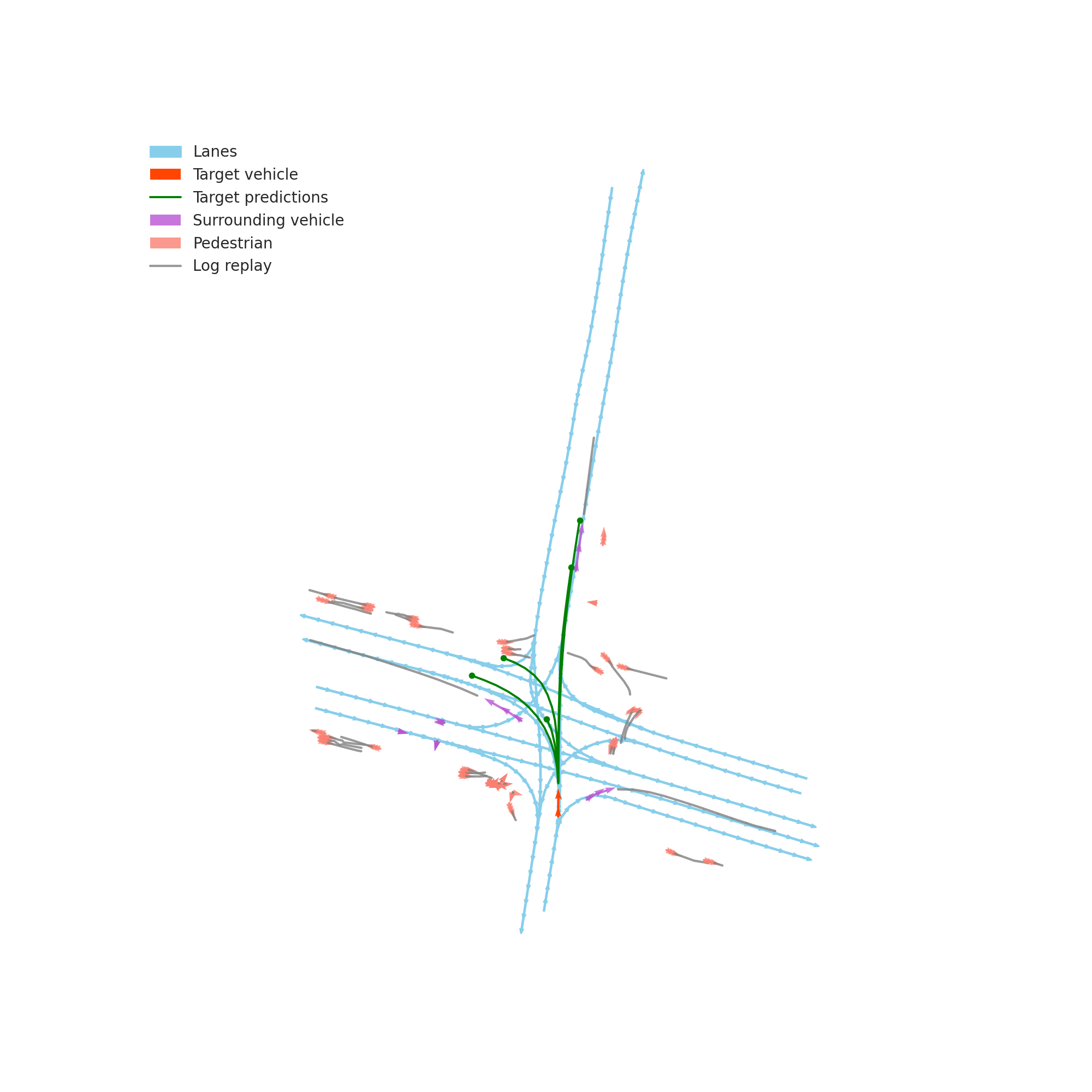}
        \end{subfigure}
        &
        \begin{subfigure}[t]{0.19\linewidth} 
            \includegraphics[width=\linewidth,trim={250 145 200 325},clip]{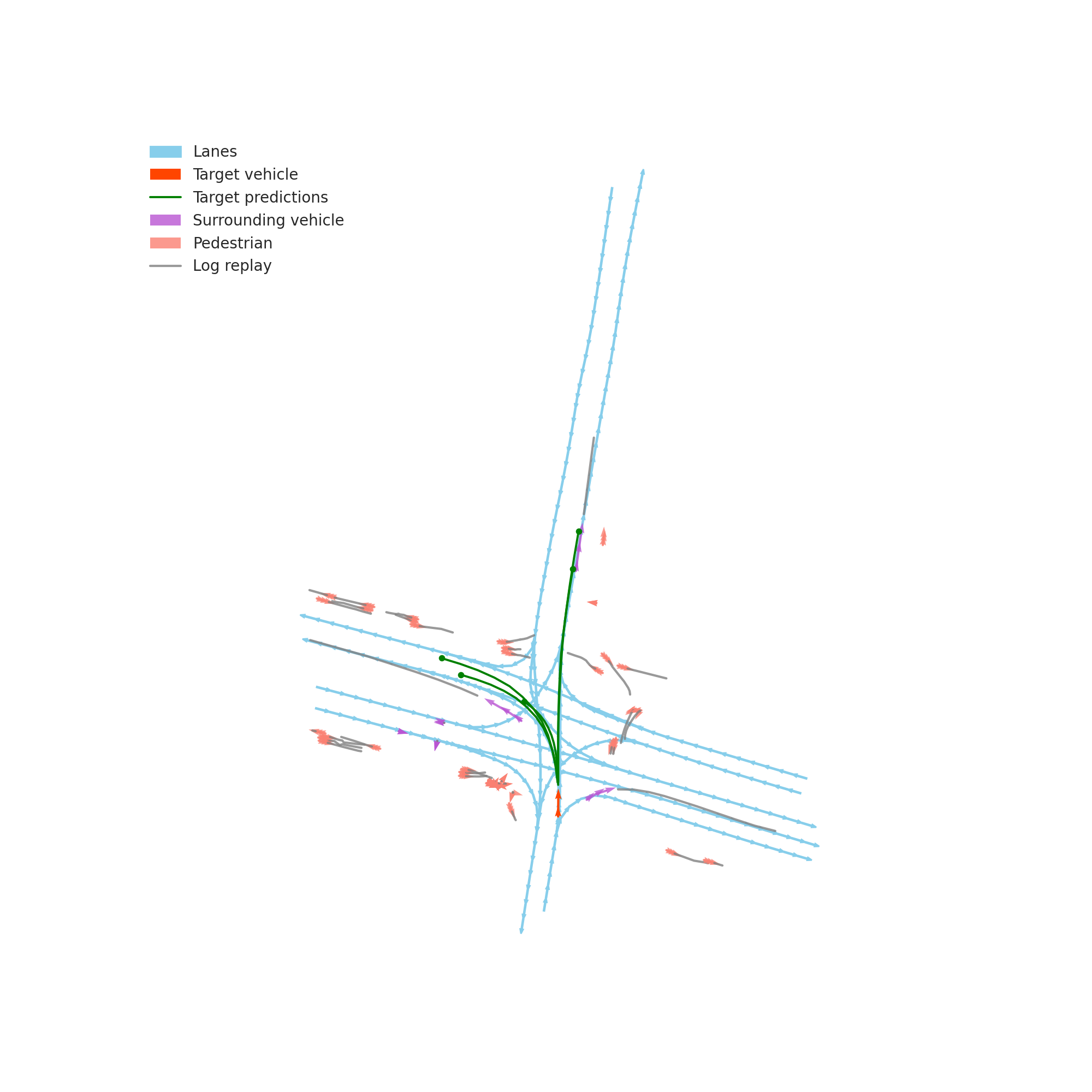}
        \end{subfigure}
        &
        \begin{subfigure}[t]{0.19\linewidth} 
            \includegraphics[width=\linewidth,trim={250 145 200 325},clip]{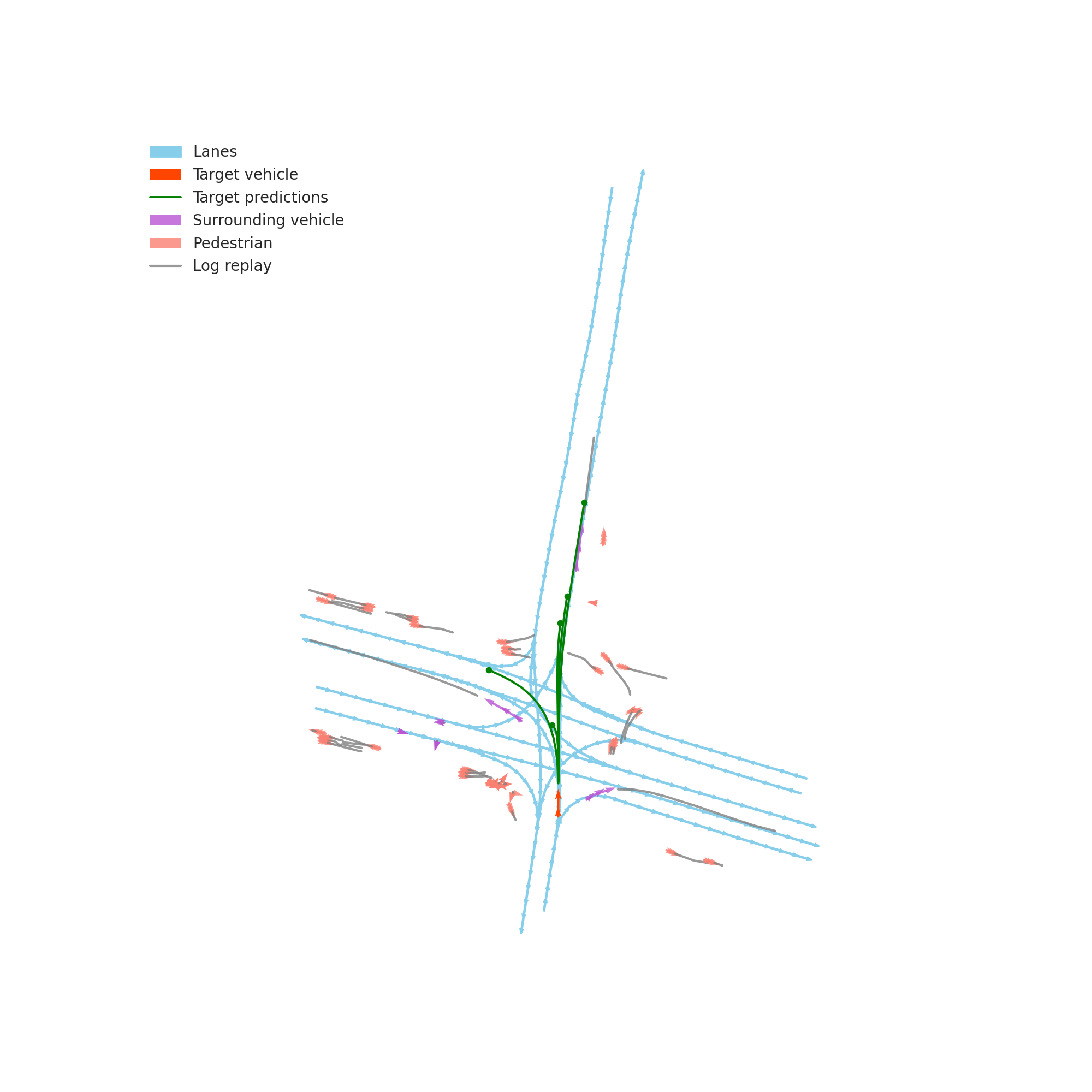}
        \end{subfigure}
        &
        \begin{subfigure}[t]{0.19\linewidth} 
            \includegraphics[width=\linewidth,trim={250 145 200 325},clip]{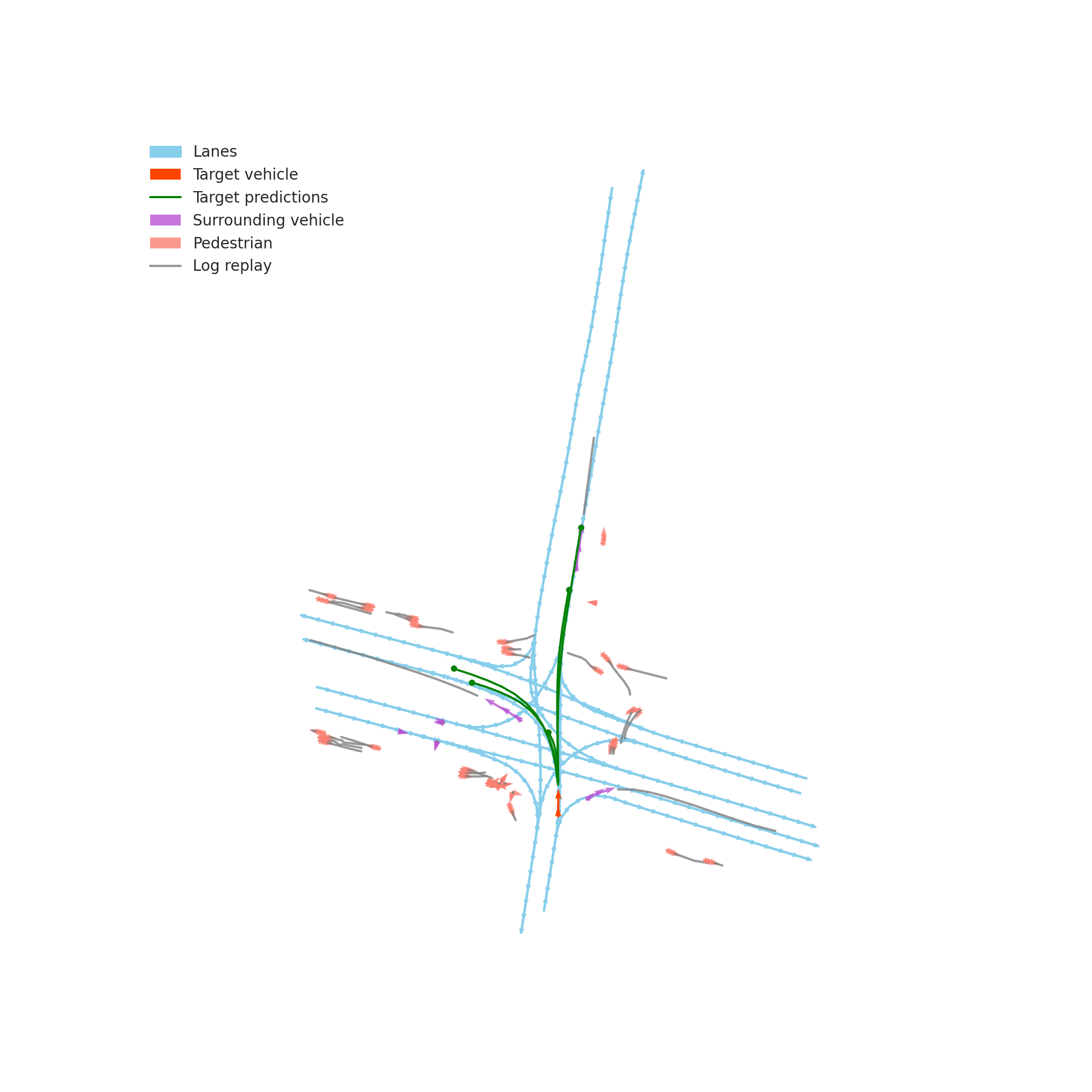}
        \end{subfigure}
        \\[-4pt]
        \hline
        \multicolumn{1}{c}{\footnotesize (a) OL, $T_{\text{sim}} = \SI{6.0}{\second}$} & 
        \multicolumn{1}{c}{\footnotesize (b) OL, $T_{\text{sim}} = \SI{2.0}{\second}$} & 
        \multicolumn{1}{c}{\footnotesize (c) CL, $T_{\text{sim}} = \SI{2.0}{\second}$} & 
        \multicolumn{1}{c}{\footnotesize (d) OL, $T_{\text{sim}} = \SI{0.5}{\second}$} & 
        \multicolumn{1}{c}{\footnotesize (e) CL, $T_{\text{sim}} = \SI{0.5}{\second}$}
        \\
    \end{tabular}
    \caption{\textbf{Qualitative comparison of OL vs. CL rollouts with different replanning periods}. While the OL model accumulates lateral drift and departs the lane, the CL model recovers lane alignment (bottom). Additionally, the CL ego decelerates for a leading vehicle during a right turn (top).}
    \label{cl_paper_fig:qual_comparision_cl_vs_OL}
    \vspace{-4mm}
\end{figure*}

Since predicting all agents at every cycle also increases computational cost with no consistent gains, we retain the log-replay setup. The absence of responsive neighbors remains a limitation, though it does not prevent the ego from responding to its surroundings: as shown in Fig.~\ref{cl_paper_fig:qual_comparision_cl_vs_OL}, a CL-trained ego both corrects for lateral drift to maintain lane alignment and decelerates for a leading vehicle during a right turn, illustrating that the model can exhibit situation-aware responses in these examples. We report these results for completeness and encourage the community to explore cost-effective reactive surrounding agents that preserve scene consistency, a step toward bringing multi-agent self-play \cite{cusumano2025robust} into the imitation learning paradigm.

\subsection{Decoder-Only Backbone and Rollout Cost}
\label{cl_paper_subsec:results_lmformer_d}

As explained in Sec. \ref{cl_paper_subsec:ablation_diff}, CL training becomes more expensive as the replanning rate increases. Table~\ref{cl_paper_tab:decoder_only} shows that removing the encoder's self- and cross-attention from LMFormer~\cite{yadav2025lmformer} yields LMFormer-D, which matches the full model on prediction quality with $45\%$ fewer parameters and $62\%$ fewer FLOPs. We present this as the design choice that makes closed-loop rollout at high replanning rates affordable, not as a contribution to open-loop accuracy.

%% file: sec/6_conclusion.tex
\section{Conclusion}
\label{cl_paper_sec:conclusion}

We presented an on-policy, closed-loop training paradigm for multimodal trajectory prediction, identifying two complementary design choices, each independently beneficial for its success: \textit{on-policy rollouts}, which preserve prediction accuracy by supervising modes exclusively from their own execution states, and \textit{gradient detachment} across state transitions, which prevents policies from coupling to a specific training replanning rate. Evaluated across three benchmarks (nuScenes, DeepScenario, \ac{av2}) and two architectures (LMFormer-D, QCNet), our approach matches open-loop baselines under single-shot evaluation and improves with replanning rate, reducing collisions by up to 35\% and off-road rates by up to 67\%. The goal-conditioned surrounding agents prediction offered no clear benefit over log-replay despite larger training costs, leaving efficient reactive scene simulation as an open challenge. We hope these insights encourage the community to view closed-loop training as a set of separable design choices, each crucial for generalization and deployment robustness.